\documentclass[10pt,twocolumn,letterpaper]{article}

\usepackage{cvpr}              %

\definecolor{cvprblue}{rgb}{0.21,0.49,0.74}
\usepackage{tcolorbox}
\usepackage[pagebackref=true,breaklinks=true,colorlinks,bookmarks=false]{hyperref}
\hypersetup{linkcolor=[rgb]{0.7,0.1,0.1}}
\hypersetup{citecolor=cvprblue}

\usepackage{tikzducks}
\usepackage{bm}
\usepackage{multirow}
\usepackage{soul}
\usepackage{wrapfig}
\usepackage{microtype}

\usepackage{caption}
\usepackage{subcaption}
\usetikzlibrary{calc,arrows,arrows.meta,backgrounds,tikzmark,shapes.geometric, decorations.pathreplacing,decorations.markings, spy,matrix,positioning}
\usepackage[ruled]{algorithm2e} %

\SetAlFnt{\small}
\SetAlCapFnt{\small}
\SetAlCapNameFnt{\small}
\SetAlCapHSkip{0pt}
\usepackage{pifont}
\usepackage{colortbl}
\usepackage{xspace}

\newlength\savewidth\newcommand\shline{\noalign{\global\savewidth\arrayrulewidth
  \global\arrayrulewidth 1pt}\hline\noalign{\global\arrayrulewidth\savewidth}}

\usepackage{listings}

\colorlet{punct}{red!60!black}
\definecolor{background}{HTML}{EEEEEE}
\definecolor{delim}{RGB}{20,105,176}
\colorlet{numb}{magenta!60!black}

\lstdefinelanguage{json}{
    basicstyle=\normalfont\ttfamily,
    numbers=left,
    numberstyle=\scriptsize,
    stepnumber=1,
    numbersep=8pt,
    showstringspaces=false,
    breaklines=true,
    frame=lines,
    backgroundcolor=\color{background},
    literate=
     *{0}{{{\color{numb}0}}}{1}
      {1}{{{\color{numb}1}}}{1}
      {2}{{{\color{numb}2}}}{1}
      {3}{{{\color{numb}3}}}{1}
      {4}{{{\color{numb}4}}}{1}
      {5}{{{\color{numb}5}}}{1}
      {6}{{{\color{numb}6}}}{1}
      {7}{{{\color{numb}7}}}{1}
      {8}{{{\color{numb}8}}}{1}
      {9}{{{\color{numb}9}}}{1}
      {:}{{{\color{punct}{:}}}}{1}
      {,}{{{\color{punct}{,}}}}{1}
      {\{}{{{\color{delim}{\{}}}}{1}
      {\}}{{{\color{delim}{\}}}}}{1}
      {[}{{{\color{delim}{[}}}}{1}
      {]}{{{\color{delim}{]}}}}{1},
}

\def\paperID{7400} %
\def\confName{CVPR}
\def\confYear{2025}

\title{\ours: Garment Estimation, Generation and Editing \\ via Large Language Models}

\newcommand{\ours}[0]{ChatGarment\xspace}
\newcommand{\oursnospace}[0]{ChatGarment}
\newcommand{\method}[0]{ChatGarment\xspace}
 
\newcommand{\supmat}{\textcolor{black}{\emph{Sup.~Mat.}}\xspace}

\author{
Siyuan Bian$^{1,2}$ \quad Chenghao Xu$^{1,3}$ \quad Yuliang Xiu$^{1,4}$ \quad Artur Grigorev$^{1,5}$ \quad Zhen Liu$^{1,6}$\\
\quad Cewu Lu$^{2}$ \quad Michael J. Black$^{1}$ \quad Yao Feng$^{7,8, \textsuperscript{\textdagger}}$ \vspace{5pt}\\
$^1$Max Planck Institute for Intelligent Systems \quad
$^2$Shanghai Jiao Tong University \\ $^3$EPFL \quad  $^4$Westlake University \quad $^5$ETH Zürich  \quad $^6$Mila, University of Montreal \\ 
\quad $^7$Meshcapade \quad $^8$Stanford University\\
\href{https://chatgarment.github.io/}{\texttt{\small chatgarment.github.io}}
}

\begin{document}

\twocolumn[{
\renewcommand\twocolumn[1][]{#1}
\maketitle
\begin{center}
    \captionsetup{type=figure}
    \includegraphics[width=\linewidth]{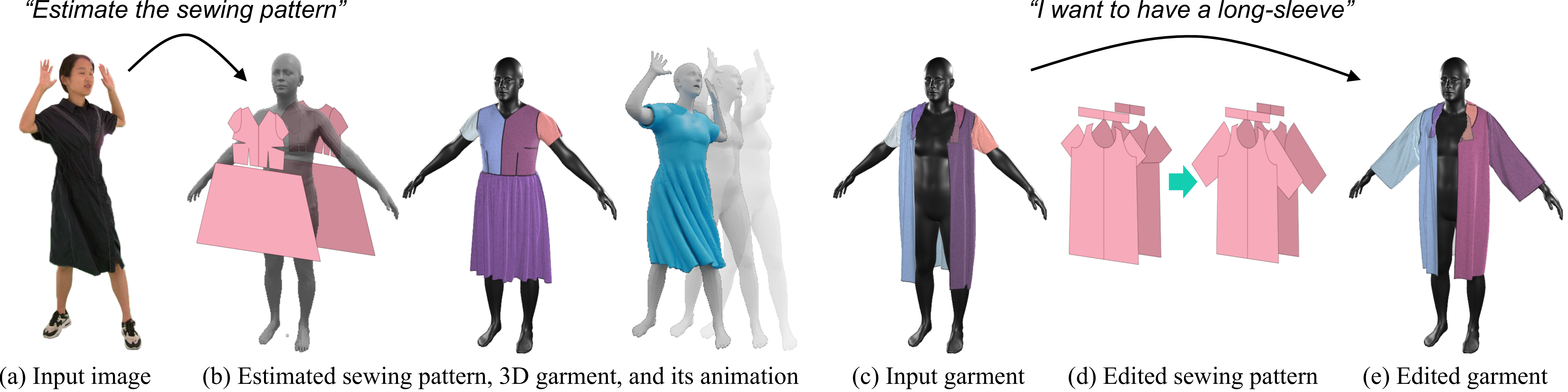}
    \captionof{figure}{\footnotesize As a multimodal 3D garment creator, \method understands both images and language. It can estimate complex 3D garments represented as sewing patterns from a single image, which can be easily animated and simulated. It also supports garment editing based on text instructions.}
    \label{fig:teaser}
\end{center} 
}]

\begin{abstract}
\def\thefootnote{\textsuperscript{\textdagger}}\footnotetext{This work was done while YF was at Meshcapade.}
We introduce ChatGarment, a novel approach that leverages large vision-language models (VLMs) to automate the estimation, generation, and editing of 3D garments from images or text descriptions. 
Unlike previous methods that struggle in real-world scenarios or lack interactive editing capabilities, ChatGarment can estimate sewing patterns from in-the-wild images or sketches, generate them from text descriptions, and edit garments based on user instructions, all within an interactive dialogue. These sewing patterns can then be draped on a 3D body and animated.
This is achieved by finetuning a VLM to directly generate a JSON file that includes both textual descriptions of garment types and styles, as well as continuous numerical attributes. This JSON file is then used to create sewing patterns through a programming parametric model. 
To support this, we refine the existing programming model, GarmentCode, by expanding its garment type coverage and simplifying its structure for efficient VLM fine-tuning. 
Additionally, we construct a large-scale dataset of image-to-sewing-pattern and text-to-sewing-pattern pairs through an automated data pipeline.
Extensive evaluations demonstrate ChatGarment’s ability to accurately reconstruct, generate, and edit garments from multimodal inputs, highlighting its potential to simplify workflows in fashion and gaming applications. 
Code and data {are} available at 
\href{https://chatgarment.github.io/}{\texttt{\small https://chatgarment.github.io/}}. 
\end{abstract}
    
\vspace{-5mm}
\section{Introduction}
\label{sec:intro} 
Imagine seeing a photo of someone in a nice looking outfit that you would like for yourself or your 3D avatar. 
Our goal is to take that photo and convert it into a 2D sewing pattern and then enable the user to use language to edit the pattern, for example, making the sleeves longer or editing the neckline. We then drape the garment on a 3D body and animate it.
This is illustrated in Fig.~\ref{fig:teaser}.
With this approach, given just an image and/or text, we construct sewing patterns for 3D garments that can be immediately used in fashion or gaming applications.
This lightweight process is in stark contrast to how sewing patterns and 3D garments are traditionally created by an artist or clothing designer.
In contrast to the traditional labor-intensive process, our approach exploits large vision-language models (VLMs) for the first time to democratize the capture and design of clothing.

Recent work explores the generation or estimation of 3D clothing using text and/or images as input \cite{jiang2020bcnet,corona2021smplicit,de2023drapenet,qiu2023rec}. 
These methods use various 3D representations including 3D meshes, point clouds, or implicit representations like Unsigned Distance Functions (UDFs) and Signed Distance Functions (SDFs). 
We observe, however, that most clothing is designed and manufactured using 2D patterns, and posit that this is a more natural representation. 
{Sewing patterns can be seamlessly integrated into existing garment design pipelines for animation and manufacturing.}
{Recent approaches~\cite{li2024isp,he2024dresscode,sewformer} use sewing patterns for garment estimation and generation, but mapping images or text to sewing patterns remains challenging due to limited training data, leading to poor generalization across diverse clothing types. These methods also does not support interactive editing, making post-processing labor-intensive for 3D artists.}
Fortunately, recent work, such as GarmentCode~\cite{GarmentCode2023,GarmentCodeData:2024}, provides a programmatic framework for encoding garment information into a JSON file. This JSON file includes textual descriptions of garment types, styles and numerical attributes, offering a rich representation of sewing patterns. The JSON file is then processed by the GarmentCode framework to produce 2D sewing patterns, which can subsequently be draped on a 3D body. 
However, to date, there is no automated method to directly extract a GarmentCode representation (JSON file) from an image or text description.

Our observation is that a JSON file resembles natural language, making it suitable for interpretation by large language models (LLMs). At the same time, Vision-LLMs (VLMs) excel at understanding images and garment concepts, such as types of garments and specific features like long sleeves.
If we enable the VLM to translate its garment understanding into the structured format of GarmentCode, then it becomes capable of generating sewing patterns. 
This allows the VLM to take an image as input and output a garment, while also supporting garment generation and editing through text descriptions or interactive dialogue. 

Specifically, we introduce \textbf{\ours}, a method that finetunes a VLM to take text queries, optionally with images, as input and outputs a JSON garment file. This JSON file is then used for GarmentCode generation. 
The JSON file includes textual descriptions of garment types (e.g., skirts, pants), styles (e.g., collar type), and continuous numerical attributes (e.g., pant length). Inspired by previous work~\cite{chatpose,golkar2023xval,kulits2024igllm}, 
{we introduce a new token (\texttt{<ENDS>}) to represent these numerical values and train an MLP projection layer to decode them from the token embedding.}
Those numbers, combined with the VLM's language output, are formatted into the final JSON file. In our approach, we keep the VLM’s vision encoder fixed, fine-tune the language model using LoRA~\cite{lora}, and jointly train the MLP projection layer. 
To adapt GarmentCode~\cite{GarmentCode2023} for more efficient VLM training, we create an improved version that supports a wider range of clothing types (e.g. open-jacket) and removes unnecessary settings to reduce token usage in training.  
We also develop an automated data construction pipeline that leverages existing tools for garment generation, draping, simulation, rendering, and automatic labeling with GPT-4o. 
This pipeline enables building a dataset with image-to-GarmentCode and text-to-GarmentCode pairs, including 20K new garments and 1M images with detailed descriptions, supporting both garment creation and editing.  

We evaluate \ours across a diverse set of tasks, including those specifically trained for, and novel applications that demonstrate the generalization capabilities of VLMs. For single-image garment reconstruction, \ours outperforms prior sewing-pattern-specific and LLM-based models on the Dress4D~\cite{wang20244d} and CloSE~\cite{antic2024close} evaluation datasets. 
Additionally, we demonstrate garment editing, generation, and multi-turn dialogue, highlighting our approach's versatility. \method can flexibly use text and images to create and edit 3D garments that can be readily animated, introducing a new workflow for garment design.

We summarize our contributions as follows: \begin{itemize}  
\item A unified model capable of estimating, generating and editing sewing patterns from multimodal inputs. 
\item The first approach to leverage VLMs for directly generating JSON files for garment creation. 
\item A refined version of GarmentCode that supports more diverse garment types and is optimized for VLM training. 
\item An automatic data construction pipeline for generating image-to-sewing-pattern and text-to-sewing-pattern data, along with the release of a large-scale dataset. 
\end{itemize}

\section{Related Work}
\label{sec:related}

\noindent\textbf{Garment Creation.} 
Most 3D garments have predefined templates (\eg, sewing patterns) and categories (\eg, skirts, shirts, pants, dresses, \etc), but their non-rigid nature makes them highly deformable with dynamic motions.
Efforts to create 3D garments can be grouped into ``template-based'' or ``free-form'' methods, based on their 3D representations.

{
A line of template-based approaches register predefined garment templates to 2D observations, which could be real captures, e.g.~from monocular video~\cite{jiang2020bcnet,qiu2023rec,luo2024garverselod,casado2022pergamo} and images~\cite{danvevrek2017deepgarment,bhatnagar2019multi,corona2021smplicit,Zhu_2022_CVPR}, generated data~\cite{shen2020garmentgeneration}, or derived from the ``proxy gradient'' computed from 2D diffusion models~\cite{Li2024GarmentDreamer3G,sarafianos2024garment3dgen} via score distillation sampling (SDS)~\cite{poole2022dreamfusion}. }
When the template is classified correctly, the predefined garment template, refined through a coarse-to-fine step, ensures topological stability and correctness.
Unfortunately, template-based methods struggle to accurately model garments that deviate from the predefined templates, particularly those with diverse styles, such as long skirts, or garments that undergo topological changes, like opening a jacket.

To address these limitations, free-form representations like (un)signed distance functions~\cite{yu2025surf,de2023drapenet,corona2021smplicit,srivastava2025wordrobe}, occupancy fields~\cite{saito2019pifu,saito2020pifuhd,xiu2022icon,xiu2022econ}, tetrahedrons~\cite{liu2024gshell,huang2024tech,shen2021dmtet}, and point clouds~\cite{POP:ICCV:2021,SkiRT:3DV:2022} are leveraged. 
However, this flexibility can compromise the inherent structure of garments, potentially leading to non-garment shapes and floating artifacts. Without shape regularization, incomplete observations lead to the reconstructed unseen parts being overly smooth. Even worse, rigging or simulating these free-form garments becomes extremely challenging.

The sewing pattern is a well-balanced solution -- with pre-defined yet flexible templates.
Several methods exploit sewing patterns to reconstruct 3D garments.
MulayCap~\cite{su2020mulaycap} estimates the parameters of a sewing pattern from a monocular video.
ISP~\cite{li2024isp} parameterizes sewing patterns as implicit functions in a 2D space. Using ISP as a garment representation, GarmentRecovery~\cite{li2024garment} trains a feedforward network to infer sewing patterns from in-the-wild photos. NeuralTailor~\cite{korosteleva2022neuraltailor} uses an RNN to produce more complex sewing patterns from point clouds. Wang \etal~\cite{sketchgarment} learn a shared garment shape space that can be inferred from multimodal input such as sketches, SewFormer~\cite{sewformer} uses a transformer to infer multi-panel patterns from monocular images, {and DressCode~\cite{he2024dresscode} estimates quantized sewing patterns with a GPT-based model from text inputs.}
Despite promising results, these approaches are limited to basic sewing patterns, featuring only front and back panels, or constrained by the limited sewing patterns seen in the training data.
Additionally, none of the above works have explored using programming parametric sewing patterns, such as GarmentCode~\cite{GarmentCode2023,GarmentCodeData:2024} to create more complex garments, either in generation or reconstruction. 

\smallskip\noindent\textbf{VLMs for 3D tasks.}
With their broad understanding of the world,
LLMs and VLMs are widely used to reason about 3D. 
There has been significant work on generating, editing, and analyzing 3D objects and scenes \cite{liu2024uni3d,xu2025pointllm,hong20233d,fu2022shapecrafter,zhang2024clay,sun20233d,hu2024scenecraft,qiu2024sgpbench}.
While these methods show the power of VLMs to reason about objects, they do not address our problem of representing 2D patterns that correspond to 3D shapes -- this is a special property of garments.
Some attempts have been made to connect sewing patterns with language, such as DressCode~\cite{he2024dresscode}. This motivates us to present \method, which connects a large language model to GarmentCode, sharing a similar approach to recent strategies~\cite{liu2023llava,chatpose} by finetuning LLMs~\cite{lora,liu2024boft,Qiu2023OFT}. 
Powered by VLMs, \method significantly improves generalization and robustness in reconstructing out-of-domain images and increases diversity in 3D clothing generation. It also supports sewing pattern editing and multi-turn dialogue, greatly simplifying the 3D garment workflow.

\section{Method}
\label{sec:method} 
\begin{figure}
    \captionsetup{type=figure}
    \includegraphics[width=\linewidth]{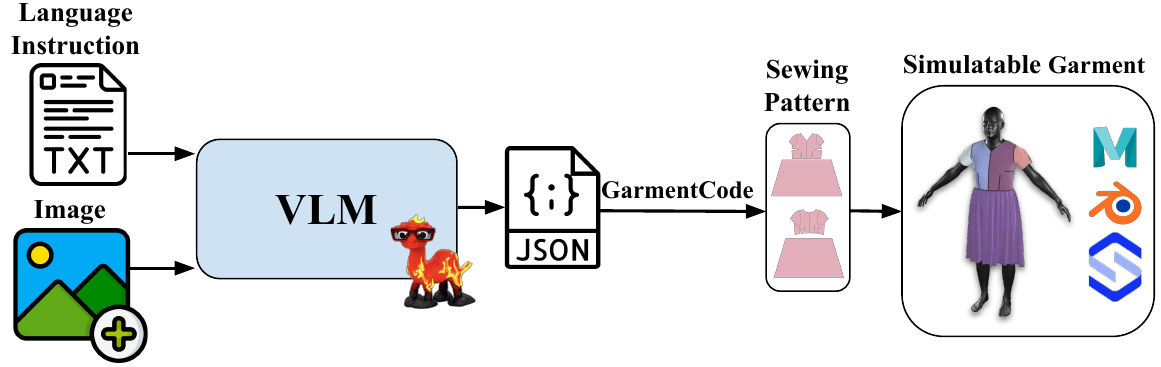}
    \captionof{figure}{\textbf{Pipeline of \method.} 
    {\method takes text or an image as input and generates a JSON file. 
    The JSON file is decoded into 2D sewing patterns using GarmentCode~\cite{GarmentCode2023} and then draped onto the human body. The final 3D garments are compatible with simulation software (\eg, MAYA, Blender, Style3D, \etc).} 
    }
    \label{fig:pipe}
    \vspace{-4mm}
\end{figure}

\subsection{Pipeline}\label{sec:pipeline}

{The goal of \ours is to simplify the workflow of garment creation by automatically estimating, generating, and editing garments from multimodal image and text input. 
To achieve this, we finetune LLaVA~\cite{liu2023llava}, a multimodal large language model $f_{\phi}$. It takes an input image $X_v$ and text prompts $X_q$, and produces a textual response $Y_t$ as $Y_t = f_{\phi}(X_q, X_v)$ or $Y_t = f_{\phi}(X_q)$ in the absence of an image. From this output, a structured JSON file $J_t$ is extracted as a subset of $Y_t$, which will be decoded into sewing patterns using the GarmentCode~\cite{GarmentCode2023} decoder $D_g$. This pipeline is illustrated in Fig.~\ref{fig:pipe}. }

\begin{figure}[!t]
    \centering
    \includegraphics[width=\linewidth]{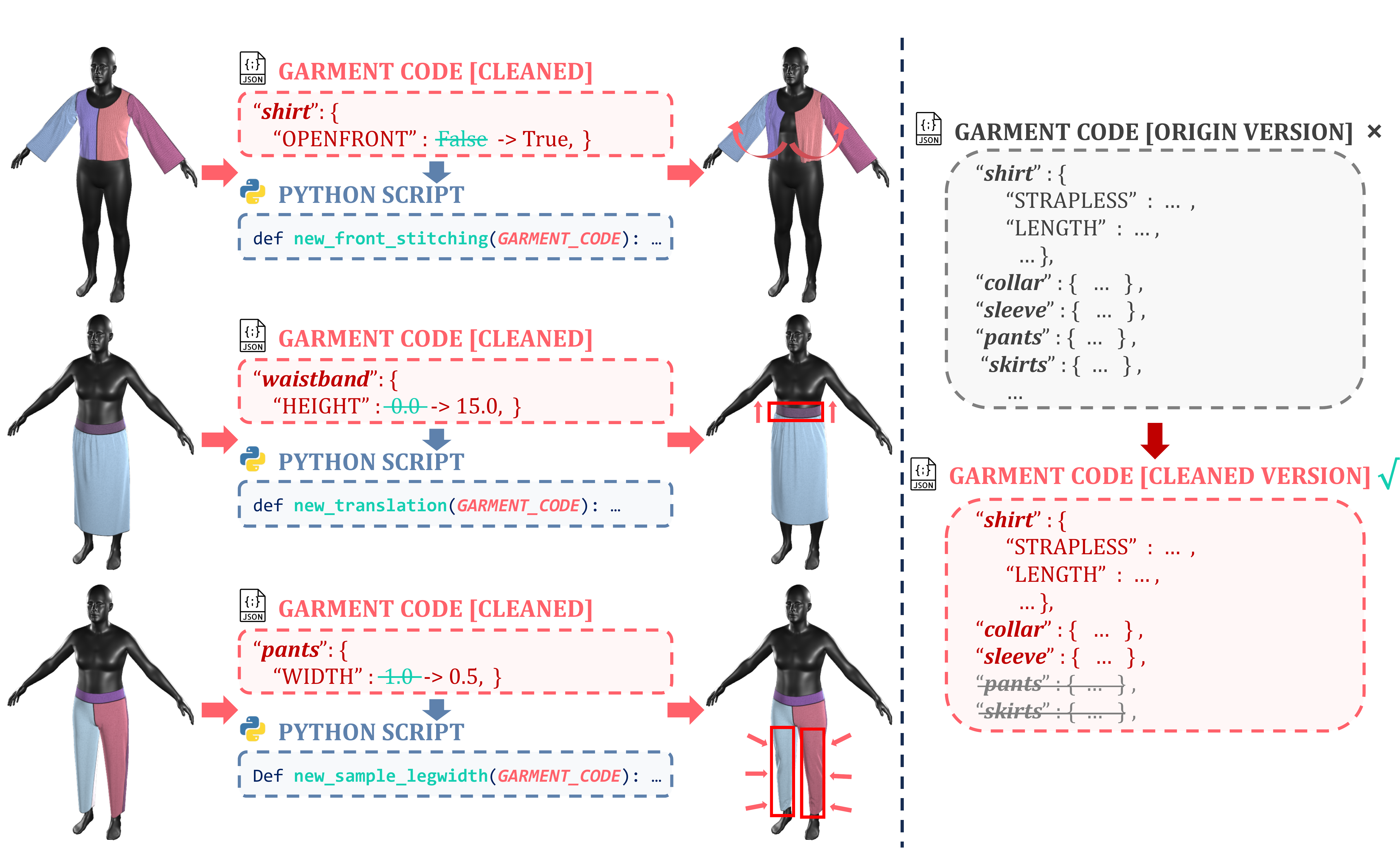}\vspace{-0.05in}
    \caption{\textbf{GarmentCodeRC}. {Left: new options to model open-front jackets, high-waist skirts, and tight pant legs. 
    Right: simplified JSON configuration for more efficient LLM training.}}
    \label{fig: method_garment}
    \vspace{-3mm}
\end{figure}

\smallskip
\noindent\textbf{LLM-friendly Rich GarmentCode.}  
We use GarmentCode to convert the JSON file into a 3D garment mesh in the canonical pose. GarmentCode~\cite{GarmentCode2023} is a programming parametric model that uses a domain-specific language (DSL) for creating 2D sewing patterns, with a hierarchical JSON structure to define garment components and stitching rules. These patterns can be draped onto body models via a warp-based pattern stitching algorithm~\cite{warp2022,GarmentCode2023}, and further be simulated easily. {GarmentCode can model diverse garments with intricate geometric details, including different cuts, frills, and pleats.}
To better align GarmentCode with our model’s language-driven requirements, we develop a refined version, GarmentCodeRC (Richer \& Cleaner) with two major improvements (see \cref{fig: method_garment}).

First, we modify stitching and panel positioning to better support diverse garment types, such as open-front jackets, high-waist skirts, and fitted pant legs. This allows us to model a broader range of real-world garments and aligns more closely with natural garment descriptions. 
Second, we simplify GarmentCode’s JSON configuration to better suit LLM training. The original configuration is redundant, applying the same settings across all garment types. We optimize this by automatically removing irrelevant settings (e.g., omitting skirt-related parameters for upper-body garments). This adjustment reduces the average language token count from 900 to 350, decreasing ambiguity in LLM training and improving efficiency. Furthermore, to stabilize model training, we normalize all floating-point values to a $[0,1]$ range. Examples of the GarmentCodeRC JSON configuration are given in \supmat
Throughout this paper, ``GarmentCode'' refers to our enriched and simplified version.

\smallskip
\noindent\textbf{LLM for Textual and Numeric Outputs.}\label{sec:llm_output}
The output JSON file $J_t$ contains both textual and numeric descriptions. Previous research~\cite{golkar2023xval,chatpose,kulits2024igllm} has shown that LLMs struggle in predicting continuous and precise numerical values. To overcome this, following ChatPose~\cite{chatpose,kulits2024igllm}, we encode these continuous values into language token embeddings and train a projection layer to recover accurate numerical values, such as garment length. 
In this setup, the language embedding for $Y_t$ is represented as $H_t$. 
From the textual output $Y_t$, upon encountering the start token \texttt{<STARTS>}, we extract the textual content of the JSON file, which spans from \texttt{<STARTS>} to \texttt{<ENDS>}. 
The language embedding at the end token \texttt{<ENDS>}, where numeric information is encoded, is represented as $H_t^{\mathbf{n}}$. A projection layer uses this embedding to predict the numerical values as $\mathbf{n} = g_{\Theta}(H_t^\mathbf{n})$. Finally, we merge these numerical values with the textual descriptions extracted earlier and format them into our final sewing pattern JSON configuration file $J_t$.

\subsection{Automatic Data Construction}
\label{sec:data_construction}
To train \ours, we need image-to-JSON and text-to-JSON paired data. While some work like SewFormer~\cite{sewformer} offers image-to-JSON pairs, they fall short in garment variety. For example, they lack tight-sleeve garments and only support a limited variety of skirts. Additionally, text descriptions for garment editing are missing in current datasets. We address these gaps by developing an automated data construction pipeline that integrates various resources.

\noindent\textbf{Garment Generation, Simulation, and Rendering.}
Following GarmentCodeData~\cite{GarmentCodeData:2024}, we sample 20,000 garments by randomizing values in sewing pattern configurations. To improve real-world garment relevance, we adjust the sampling ratios and assign lower weights for asymmetric garments and those with intricate cuffs. To create outfits, we pair top and bottom garments together or use single-piece items like dresses or jumpsuits. The garment set is draped on a SMPL-X~\cite{smplx} neutral model.

\begin{figure}
    \captionsetup{type=figure}
    \centering
    \includegraphics[width=\linewidth]{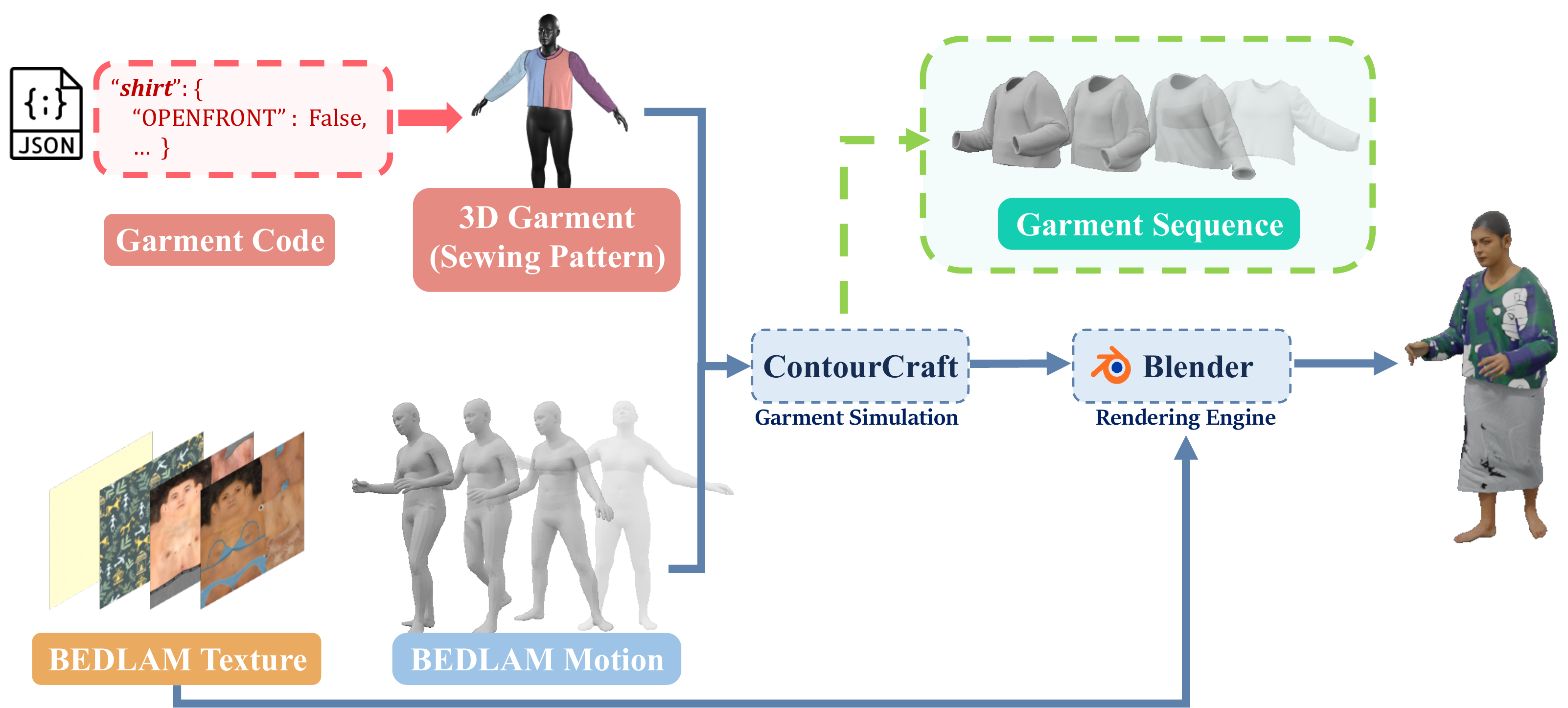}
        \captionof{figure}{\textbf{Data Construction Pipeline.} We generate garments from JSON configurations, simulate them with ContourCraft~\cite{contourcraft} and render with Blender.}
    \label{fig:simulation_pipeline}
    \vspace{-3mm}
\end{figure} 

{We automate garment simulation using ContourCraft~\cite{contourcraft}, chosen for its stability and scalability, with SMPL-X motion sequences sampled from the BEDLAM dataset~\cite{bedlam}. 
Specific garment vertices are attached to corresponding body vertices using predefined rules to prevent garments from slipping off. For example, the upper edges of pants and skirts are connected to the nearest body vertices, and similar attachments are made for upper-body garment vertices near the shoulders. 
We filter out simulations with significant interpenetration or improper positioning in two steps.}
First, GarmentCode performs a sewing-pattern validation check, ensuring the integrity of panel edges and stitching. Second, we verify simulation results by assessing inter-penetration, and identify garments at risk of slipping off by inspecting the proximity of each garment vertex to the nearest body part.
Each garment is then rendered in Blender from four random perspectives at elevation angles between 0 and 30 degrees, with garment and body textures sourced from the BEDLAM dataset. These steps are shown in Fig.~\ref{fig:simulation_pipeline}.

\smallskip
\noindent\textbf{GPT-4o Labeling.}\label{subsec:dataset}
The previous method~\cite{he2024dresscode} prompts GPT-4o to generate the general geometric descriptions for garments, such as the garment types, lengths, widths and sleeve lengths. However, these general descriptions are insufficient for garment detail understanding: the finetuned VLM struggles to interpret detailed features of garment components, such as pant cuffs, or execute specific garment edits. To address this, we construct a new dataset with low-level descriptions for individual garment parts.
For image-based reconstruction, we instruct GPT-4o to generate descriptions for all visible garment parts in the image. For garment editing, we sample 5,000 garment parts (shirt main body, collar, sleeves, sleeve cuffs, waistband, pant legs, pant cuffs, and skirts), and use GPT-4o to generate descriptive text for each part. Assembling these components yields 20,000 new garments with detailed annotations. 

For garment editing, we randomly select two garments that differ in certain parts, with the editing prompt being: ``\texttt{Change the garment sewing pattern by modifying <PART-1> to <DESC-1>, <PART-2> to <DESC-2>, $\cdots$ while keeping other parts unchanged.}''. \texttt{<PART-1>, <PART-2>,} $\cdots$ represent part names that differ between the two garments, and \texttt{<DESC-1>, <DESC-2>} are target garment part descriptions.

We finally build a synthetic dataset of 40K garments and 1 million images with text annotations, supporting garment reconstruction and editing from multimodal inputs. Plus, we use GPT-4o to label part-level details of 38K SHHQ~\cite{fu2022styleganhuman} images to make the VLM better understand in-the-wild garment images. 
Compared to previous garment sewing pattern datasets~\cite{sewformer}, our dataset offers:
1) \textbf{Diversity}: GarmentCode generates a wider range of real-world garment types; 
2) \textbf{Detailed Descriptions}: Garments are labeled with detailed geometric features; 
3) \textbf{Precise Control}: Allows for precise, low-level garment manipulation.

\begin{figure*}[!th]
\centering
    \captionsetup{type=figure}
    \includegraphics[width=0.9\linewidth]{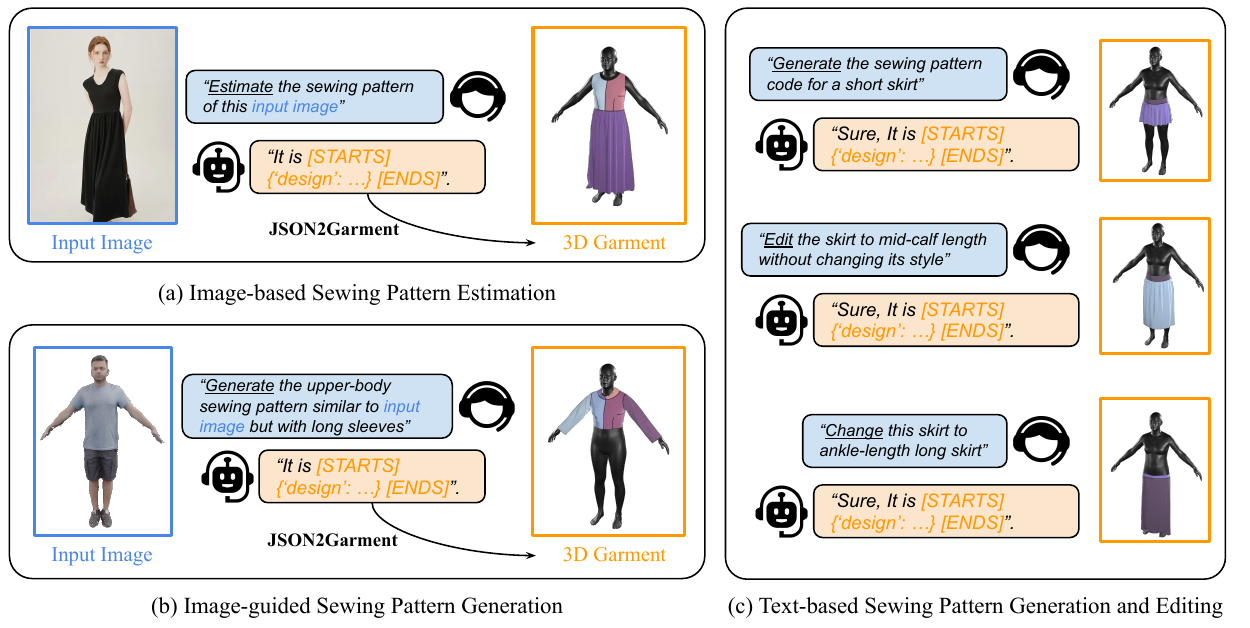}  \vspace{-0.1in}
    \captionof{figure}{{\textbf{ChatGarment's dialog modes.} Images and texts are adaptively combined to guide garment generation and editing. Text output between \texttt{{STARTS}} and \texttt{{ENDS}} contains information about the JSON configuration, which is then converted to 3D garments ({JSON2Garment}) for visualization and simulation purposes.}}
    \label{fig:dialog}
    \vspace{-3mm}
\end{figure*} 

\subsection{Training}\label{sec:training}
{We use LLaVA~\cite{liu2023llava} as our base VLM and finetune it with LoRA~\cite{lora}, with its parameters denoted as $\phi_{lora}$. Additionally, we optimize the LLM token embedding $e_{token}$, the LLM head layer \texttt{lm\_head}$_{\Theta'}$, and the projection layer $g_{\Theta}$.}

\smallskip
\noindent\textbf{Loss Function.}
As described in Sec.~\ref{sec:llm_output}, the numeric embedding is extracted from the token \texttt{[ENDS]} as $H_t^\mathbf{n}$. An MLP projection layer maps the embedding $H_t^\mathbf{n}$ to 76 float values $\mathbf{n}$ in the JSON configuration. 
{
Given the ground truth text output $\hat{Y}_t$ and sewing pattern values $\hat{\mathbf{n}}$, we optimize with:
\begin{equation}
    \mathcal{L} =  \text{CE}(\hat{Y}_t, Y_t) + \lambda_\mathbf{n} \left|(\hat{\mathbf{n}} - \mathbf{n}) \cdot \mathbf{m} \right|,
\end{equation}
where the first term is a cross-entropy loss, and the second term is the L1 difference for numeric values. $\mathbf{m}$ is a mask that filters out irrelevant numeric values, and $\lambda_\mathbf{n} = 0.1$ is the weight for the L1 loss. See more training details in \supmat
}

\subsection{Multi-turn Dialogue with Multimodal Inputs.}
\label{sec:applications}
After training, our model can handle multimodal input reconstruction, garment generation, and editing, as reflected in the training data. Additionally, it exhibits \textit{zero-shot} reasoning for garment sewing patterns within multi-turn dialogues, despite being trained only on single-turn dialogue datasets. 
This enables users to create a garment from an image or text description and then refine it within the same interaction.  
A typical workflow might start with the VLM estimating a sewing pattern from an image. After reviewing the result, the user can request specific adjustments, providing a flexible, user-centered design process (see Fig.~\ref{fig:dialog} for an example).

\section{Experiments}
{We conduct experiments on image-based reconstruction (Sec~\ref{sec:img_recon}), garment editing (Sec~\ref{sec:img_edit}) and text-based garment generation (Sec~\ref{sec:text_gen}), along with ablation studies (Sec~\ref{sec:ablation}).}

\subsection{Image-based Reconstruction}\label{sec:img_recon}
\noindent\textbf{Benchmarks}.
We conduct image-based reconstruction experiments on the CloSe \cite{antic2024close} and Dress4D \cite{wang20244d} datasets. CloSe is a large-scale scan collection including 3D clothing segmentations across 18 clothing categories. 
Following CloSe, we use 145 scans with accurate SMPL-X fits as our validation set. 
As CloSe offers limited instances of loose garments, such as skirts and dresses, we further add samples from Dress4D, which has real-world outfits recorded using a multi-view volumetric capture system. We include 4 loose-fitting outfits and 36 images rendered from them. 

\smallskip
\noindent\textbf{Implementation Details.}
Our approach applies the Chain-of-Thought (CoT)~\cite{wei2022chain} method. {We first prompt ChatGarment to generate a text description of the outfit in the image. The generated text descriptions are combined with the input image to estimate the final garment JSON configuration.}
More details are provided in \supmat

\smallskip
\noindent\textbf{Baselines}.
We compare our method with two previous approaches, SewFormer~\cite{sewformer} and GarmentRecovery~\cite{li2024garment}. We use the publicly available code and checkpoints for these methods in our evaluation. Additionally, we introduce several new baseline models for comparison: 

\begin{itemize}
    \item  {\textbf{DressCode}. We employ DressCode~\cite{he2024dresscode} to generate garments from image descriptions. Specifically, we utilize GPT-4o to generate garment descriptions from input images. These descriptions are processed by DressCode to create sewing patterns.}
    \item \textbf{GPT-4o}. We design prompt templates and query GPT-4o using in-context learning (ICL)~\cite{dong2022survey}. We use Garment Generator~\cite{korosteleva2021generating} as the sewing pattern configuration tool, as it has significantly fewer possible configurations and allows for more efficient template design and significantly fewer conversation rounds. We first prompt GPT-4o to choose the most suitable template from a predefined list, then query the parameter value by displaying three garment renderings with uniformly sampled parameter values.
    \item \textbf{LLaVA*}. In this setup, we use the LLaVA model to directly output the numerical values for the sewing pattern instead of using the sewing pattern projection layer. The remaining components are identical to our method.
    \item \textbf{LLaVA-T}. We first employ GPT-4o to generate detailed textual descriptions of the garment, and \ours is prompted to reconstruct the sewing pattern based solely on the text description. 
    \item \textbf{LLaVA-I}. We input the image into the model and prompt it to directly output the sewing pattern without referring to text descriptions.
\end{itemize}

\begin{table}
\scriptsize
\setlength{\tabcolsep}{3pt}
\renewcommand{\arraystretch}{1.3}
\centering
	\centering
     \resizebox{\columnwidth}{!}{
    \begin{tabular}{l|cc|cc|c}
       \multirow{2}{*}{Methods} & \multicolumn{2}{c|}{Target-Pose} & \multicolumn{2}{c|}{A-Pose} & \\
        &  CD ($\downarrow$) & F-Score ($\uparrow$) & CD ($\downarrow$) & F-Score ($\uparrow$) & Failure Rate ($\downarrow$) \\
        \shline
        Sewformer~\cite{sewformer} & 27.06 & 0.55 & 20.66 & 0.58 & 4.3\%\\
        DressCode~\cite{he2024dresscode} & 20.16 & 0.60 & 12.12 & 0.61 & 11.4\%\\
        {GarmentRecovery~\cite{li2024garment}} & 9.69 & 0.55 & 5.75 & 0.64 & - \\
        GPT-4o & 11.04 & 0.62 & 10.44 & 0.65 & 0 \\
        \hline
        LLaVA* & 5.29 & 0.72 & 6.29 & 0.76 & 0 \\
        LLaVA-T & 6.17 & 0.72 & 7.25 & 0.74 & 0  \\
        LLaVA-I & 4.32 & \textbf{0.75} & 4.93 & \textbf{0.78} & 0 \\
        Ours & \textbf{3.12} & \textbf{0.75} & \textbf{3.06} & \textbf{0.78} & 0 \\
    \end{tabular}}
    \vspace{-2mm}
\caption{\textbf{Quantitative comparisons of Image-based garment reconstruction on Dress4D.} \method achieves the best performance on all metrics.}
\label{table:recon-image-dress4d}
\vspace{-5mm}
\end{table}

\begin{table}
\scriptsize
\setlength{\tabcolsep}{3pt}
\renewcommand{\arraystretch}{1.3}
\centering
	\centering
     \resizebox{0.85\columnwidth}{!}{
    \begin{tabular}{l|ccc}
       \multirow{2}{*}{Methods} & \multicolumn{3}{c}{CLoSE} \\
         & CD ($\downarrow$) & F-Score ($\uparrow$) & Failure Rate ($\downarrow$) \\
        \shline
        SewFormer~\cite{sewformer} & 9.70 & 0.708 & 8.33 \%\\
        DressCode~\cite{he2024dresscode} & 15.77 & 0.616 & 7.0\% \\
        {GarmentRecovery~\cite{li2024garment}} & \textbf{2.39} & 0.785 & - \\
        GPT-4o & 3.52 & 0.755 & 0 \\
        \hline
        LLaVA* & 3.08 & 0.775 & 0 \\
        LLaVA-T & 3.76 & 0.779 & 0 \\
        LLaVA-I  & 4.55 & 0.761 & 0 \\
        Ours & 2.94 & \textbf{0.790} & 0 \\
    \end{tabular}}
\vspace{-2mm}
\caption{\textbf{Quantitative comparisons of Image-based reconstruction on CloSe.} The garments are deformed to the target pose. 
}
\label{table:recon-image-close}
\vspace{-4mm}
\end{table}

{
Since methods use different default human poses, we apply Linear Blend Skinning~\cite{abdrashitov2023robust} to re-pose all garments to the ground-truth body shape and pose for fair comparison.
}

\begin{figure*}[!ht]
    \centerline{    \includegraphics[width=\linewidth]{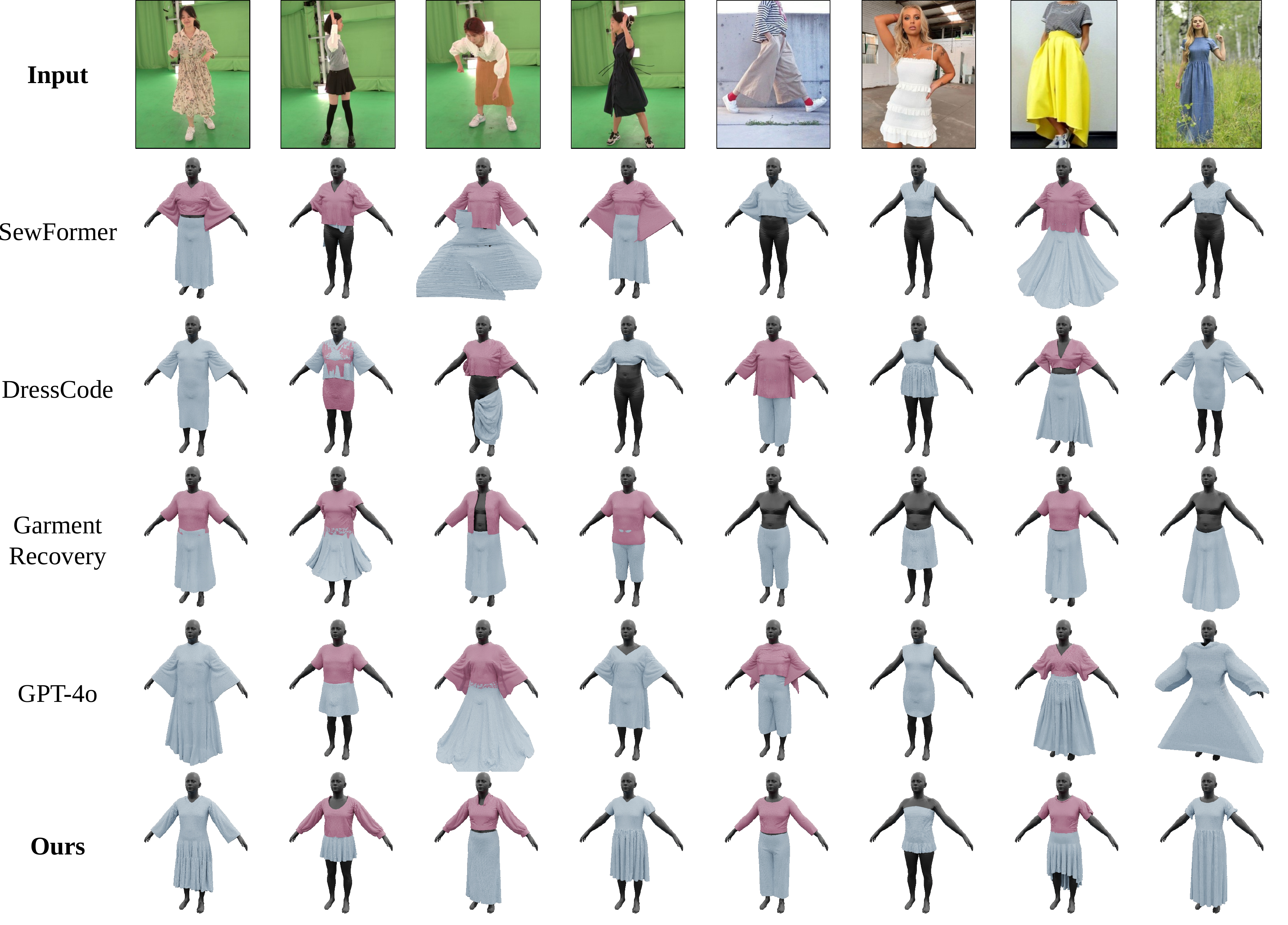}}
    \vspace{-0.05in}
       \caption{ \textbf{Image-based Garment Reconstruction.} Unlike SewFormer~\cite{sewformer}, DressCode~\cite{he2024dresscode}, GarmentRecovery~\cite{li2024garment}, and the GPT-4o-based method, which often make mistakes or ignores garments. \method faithfully captures the shape, style and composition of the garments. 
       }
    \label{fig:imgrecon}
    \vspace{-4mm}
\end{figure*}

\smallskip
\noindent\textbf{Results.}
We employ the mean chamfer distance, the mean F-Score at $\tau = 0.001$ and the stitching failure rate as quantitative metrics. 
As~\cref{table:recon-image-dress4d} and ~\cref{table:recon-image-close} show, \ours significantly outperforms existing methods on loose garments in Dress4D. 
On the CloSE dataset, which mostly features tight garments, GarmentRecovery~\cite{li2024garment} does well by leveraging garment segmentation and normal maps as guidance in an optimization-based framework. However, these cues are insufficient for accurately reconstructing loose garments. 
Our method has similar performance in this case, but is more general.
Also, our algorithm robustly reconstructs garments with no stitching failures. In contrast, DressCode and SewFormer often generate invalid garment panels leading to failed stitching. The primary difference lies in the output format: DressCode and SewFormer predict the panel edges and stitches of the sewing pattern, while our method and the GPT-4o-based approach estimate the JSON configuration. {Although direct edge and stitch prediction offers greater flexibility in theory, our experiments show that it does not enhance reconstruction accuracy. Instead, it increases the likelihood of generating invalid sewing patterns when handling out-of-distribution (OOD) data.}

Our model demonstrates superior performance compared to LLaVA*, highlighting the effectiveness of the proposed projection layer. Furthermore, our CoT-based two-step reconstruction algorithm outperforms both image-based (LLaVA-I) and text-based (LLaVA-T) reconstruction baselines. 
{The text descriptions detail garment types and styles, offering valuable information for reconstruction, while images reveal intricate details that are hard to describe in words. Thus, combining both text and image inputs is the most effective strategy for image-based garment reconstruction.
}

\Cref{fig:imgrecon} shows our model accurately estimates garment type, style, and details like skirt hems and collar styles. In contrast, other methods often misinterpret or overlook these details. 
{SewFormer occasionally fails to reconstruct garments due to detection or stitching failure, and GarmentRecovery lacks support for whole-body dress reconstruction. Additionally, inaccuracies in 2D garment segmentation~\cite{li2020self} can lead to errors in GarmentRecovery reconstruction.}

\begin{figure*}[!t]
    \centering
    \includegraphics[width=\linewidth]{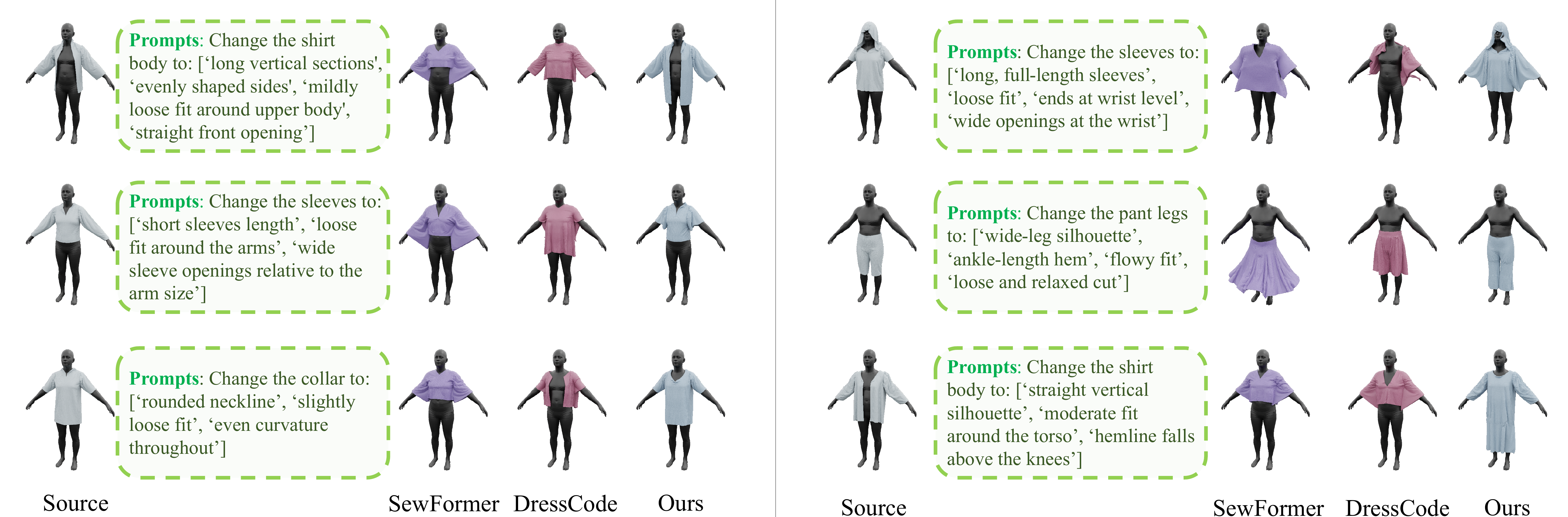}
       \caption{
            \textbf{Results on Garment Editing.} We present six examples where the models (SewFormer~\cite{sewformer}, DressCode~\cite{he2024dresscode}, and \ours) edit the source garment shown in the source image according to the given editing instructions in the prompt boxes.
        } 
    \label{fig:editing}
\vspace{-4mm}
\end{figure*}

\begin{figure}[!t]
    \captionsetup{type=figure}
    \centering
    \includegraphics[width=0.9\linewidth]{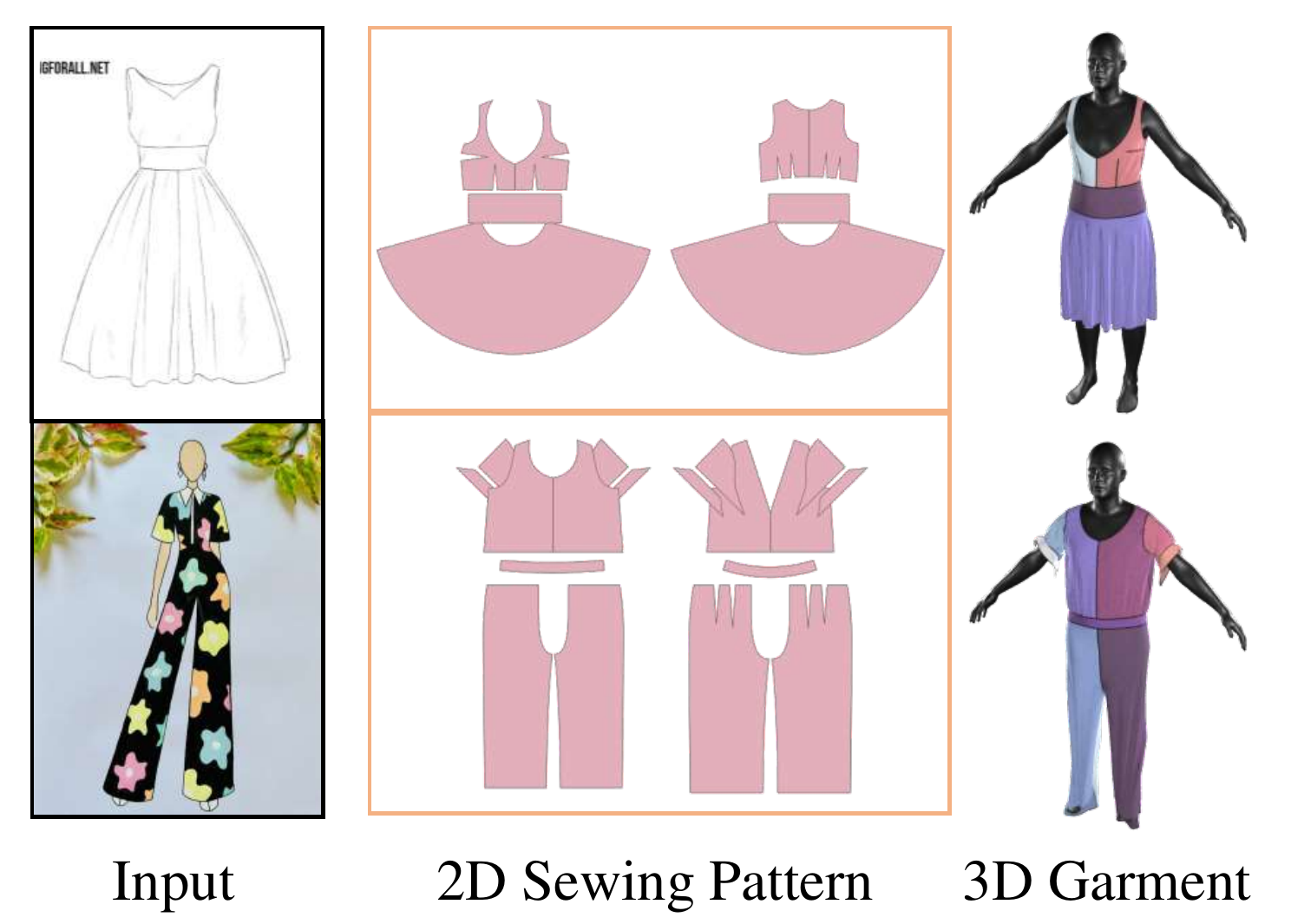}
    \captionof{figure}{\textbf{Reconstruction results from garment sketches.} The generated garments accurately capture the shape and style of the input sketch, despite our model being trained without sketches.} 
    \label{fig: sketch}
\vspace{-4mm}
\end{figure} 

Garment designers often create image sketches before making sewing patterns. Surprisingly, despite not being trained on garment sketch images, \ours demonstrates strong generalization, accurately predicting garment types and styles from sketches (\cref{fig: sketch}).

\subsection{Garment Editing}\label{sec:img_edit}
\noindent\textbf{Benchmarks.} 
To evaluate garment editing performance, we construct an extra evaluation dataset using methods outlined in Sec.~\ref{subsec:dataset}. We manually select garment pairs with clear textual descriptions and substantial visual differences. This dataset consists of 135 garment pairs, each accompanied by corresponding images and text descriptions.
The evaluation of garment editing involves three components: the source garment, the target garment, and the editing prompts. The goal is to derive the sewing pattern for the target garment using the source garment and the editing prompts. 

\smallskip
\noindent\textbf{Implementation Details.}
For garment editing, we concatenate the source garment JSON file and the editing prompt together and pass it into \ours.

\smallskip
\noindent\textbf{Baselines.}
We modify DressCode~\cite{he2024dresscode} and SewFormer~\cite{sewformer} for comparison with \ours:
\begin{itemize}
    \item {\textbf{DressCode with edited text.} We generate a text description of the edited garment and use DressCode to create the corresponding sewing patterns. The description is produced by GPT-4o, which takes as inputs: 1) garment editing instructions and 2) front and back views of the source garment. The generated description follows DressCode's original input format.}
    \item \textbf{SewFormer with edited images.} We obtain the front view of the edited garment with TurboEdit~\cite{deutch2024turboedit} to which we pass 1) the front view of the source garment, 2) the text description of the source garment, and 3) the editing instructions, and then use SewFormer to turn the image of the edited garment into the corresponding sewing pattern. 
\end{itemize}

\smallskip
\noindent\textbf{Results.}
We compare our method with SewFormer and Dresscode on Chamfer Distance and F-score in \cref{table:garment_edit}, where our approach demonstrates superior performance. The qualitative results in Fig.~\ref{fig:editing} show that our method accurately modifies the targeted garment parts according to the input prompt. In contrast, SewFormer and DressCode often fail to precisely follow the prompt instructions and frequently alter other unintended areas of the garment.

\subsection{Text-based Generation}
\label{sec:text_gen}
Following DressCode~\cite{he2024dresscode}, we use GPT-4o to generate 150 highly diverse text labels from 85 selected in-the-wild images featuring various tops, skirts, pants, and dresses. Next, we render the generated 3D garments with default textures and compute the CLIP score using the provided text labels.
See quantitative results in~\cref{table:text_gen}. Our model achieves a higher CLIP score compared to DressCode~\cite{he2024dresscode}, highlighting the superior prompt-following capabilities of our approach. The qualitative results are shown in \supmat

\begin{table}[!t]
\renewcommand{\arraystretch}{1}
\centering
\footnotesize
    \begin{tabular}{c|cc}
       {Methods} & {CD ($\downarrow$)} & F-Score ($\uparrow$) \\
        \shline
        SewFormer~\cite{sewformer} & 28.77 & 0.548 \\
        DressCode~\cite{he2024dresscode} & 10.97 & 0.750 \\
        Ours  & \textbf{2.51} & \textbf{0.893} \\
    \end{tabular}
\vspace{-2.0mm}
\caption{\small \textbf{Quantitative comparisons for garment editing.} \method outperforms other two methods by a large margin.}
    \label{table:garment_edit}
\vspace{-1.0mm}
\end{table}

\begin{table}[!t]
\renewcommand{\arraystretch}{1}
\centering
\footnotesize
    \begin{tabular}{c|c}
       {Methods} & {CLIP score} ($\uparrow$) \\
        \hline
        DressCode~\cite{he2024dresscode} & 22.8 \\
        Ours  & \textbf{23.7}  \\
    \end{tabular}
\vspace{-2.0mm}
\caption{\small \textbf{Quantitative comparisons for text-based garment generation.} \method achieves a higher CLIP score than DressCode, indicating better alignment with the text inputs.}\label{table:text_gen}
\vspace{-3.0mm}
\end{table}

\subsection{Ablation}\label{sec:ablation}
We evaluate the impact of different aspects of \ours, including multi-modal LLM backbones, and various training datasets. Please refer to \supmat

\section{Conclusion} 
We introduce \ours, a VLM-based model that estimates garments from images and generates or edits garments based on text descriptions. Leveraging the capabilities of LLMs, \ours supports multi-turn dialogue interactions for 3D garment design. It surpasses existing sewing pattern-specific and LLM-based methods in accurately estimating sewing patterns from a single image.  
While \ours excels in many areas, it may struggle with precise garment editing. For instance, altering the skirt length might unintentionally affect other parts of the garment. This issue could be addressed with more detailed text instructions or a differential CSG optimizer~\cite{yuan2024diffcsg} for finer adjustments. 
Nonetheless, \ours offers the first VLM-based framework for unified garment estimation, generation, and editing. Future directions could include expanding into garment manufacturing design or improving clothing simulation with more fine-grained material attributes.

\clearpage
\clearpage

\section*{Acknowledgments and Disclosure} 

We thank \textit{Maria Korosteleva} for help on GarmentCode, and  \textit{Giorgio Becherini} for advice on data acquisition and processing.
\textit{Yuliang Xiu} is funded by Max Planck Institute for Intelligent Systems, the Research Center for Industries of the Future (RCIF) at Westlake University, and the Westlake Education Foundation. 

While MJB is a co-founder and Chief Scientist at Meshcapade, his research in this project was performed solely at, and funded solely by, the Max Planck Society.

{
    \small
    \bibliographystyle{ieeenat_fullname}
    \bibliography{main}
}

\maketitlesupplementary
\appendix

\renewcommand{\thesection}{S\arabic{section}}
\renewcommand{\thefigure}{S\arabic{figure}}
\renewcommand{\thetable}{S\arabic{table}}
\setcounter{section}{0}
\setcounter{figure}{0}
\setcounter{table}{0}

\noindent In the supplementary material, we include additional details on training and evaluation, as well as ablation studies and qualitative visualizations.

\section{More Implementation Details}

\subsection{GarmentCodeRC}

\noindent\textbf{Garment Sewing Pattern.}
We improve the JSON-format sewing pattern configurations provided by GarmentCode~\cite{GarmentCode2023,GarmentCodeData:2024}. The original GarmentCode JSON configuration is a fixed-length file containing the same entries for all garments. We optimize this by adding new features, automatically removing irrelevant settings during garment construction (\eg, omitting skirt-related parameters for upper-body garments) and normalizing floating-point values to $[0,1]$. 
Here are two GarmentCodeRC JSON files for a skirt and a pair of pants:
\begin{lstlisting}[language=json,firstnumber=1,basicstyle=\ttfamily\footnotesize]
{
    "meta": {
        "upper": "None",
        "wb": "FittedWB",
        "bottom": "PencilSkirt"
    },
    "waistband": {
        "waist" 0.501,
        "width": 0.205,
        "height": 5
    }
    "pencil-skirt": {
        "length": 0.365,
        "rise": 0.988,
        "flare": 0.577,
        "low_angle": 5,
        "front_slit": 0.010,
        "back_slit" 0.009,
        "left_slit": 0.001,
        "right_slit": 0.001,
        "style_side_cut": "Sun"
    }
} %
\end{lstlisting}

\begin{lstlisting}[language=json,firstnumber=1,basicstyle=\ttfamily\footnotesize]
{
    "meta": {
        "upper": "None",
        "wb": "None",
        "bottom": "Pants"
    },
    "pants": {
        "length": 0.203,
        "width": 0.062,
        "flare": 0.516,
        "rise": 0.816,
        "cuff": {"type": "None"}
    }
} %
\end{lstlisting}

\noindent\textbf{Outfit Sewing Pattern.}
From real images, we see that people often wear multiple garments, like a T-shirt and pants. In these cases, we combine them into a single outfit represented as a new JSON dictionary. If the subject wears one upper and one lower garment, the model will return  in this format:

\begin{lstlisting}[language=json,firstnumber=1,basicstyle=\ttfamily\footnotesize]
{
    "upperbody_garment": {
        (upper garment sewing pattern)
    },
    "lowerbody_garment": {
        (lower garment sewing pattern)
    }
}
\end{lstlisting}

Otherwise, if the subject wears a single whole-body garment (\eg, dresses or jumpsuits), the model will return in the following format:

\begin{lstlisting}[language=json,firstnumber=1,basicstyle=\ttfamily\footnotesize]
{
    "wholebody_garment": {
        (wholebody garment sewing pattern)
    }
}
\end{lstlisting}

\subsection{\ours Training Data}
{
\noindent\textbf{Training Data Overview.}
The training data consists of four parts: garment reconstruction data (35\%), garment description data (15\%), garment editing data (15\%), and visual instruction tuning data (35\%). }
{\begin{itemize}
    \item \textbf{Garment Reconstruction Data}: Includes 20,000 simulated garments with images rendered by Blender and text labels generated by GPT-4o. During training, text labels and images are omitted from the input with a 25\% probability respectively.
    \item \textbf{Garment Description Data}: Contains 38,000 SHHQ images with text descriptions generated by GPT-4o.
    \item \textbf{Garment Editing Data}: Comprises 20,000 garments generated following the rules in section~\ref{subsec:dataset}.
    \item \textbf{Visual Instruction Tuning Data}: Utilizes LLaVA-v1.5-mix665k dataset\footnote{\href{https://huggingface.co/datasets/liuhaotian/LLaVA-Instruct-150K}{liuhaotian/LLaVA-Instruct-150K}}.
\end{itemize}}

\begin{figure*}
\vspace{-5pt}
    \captionsetup{type=figure}
    \centering
    \includegraphics[width=0.9\linewidth]{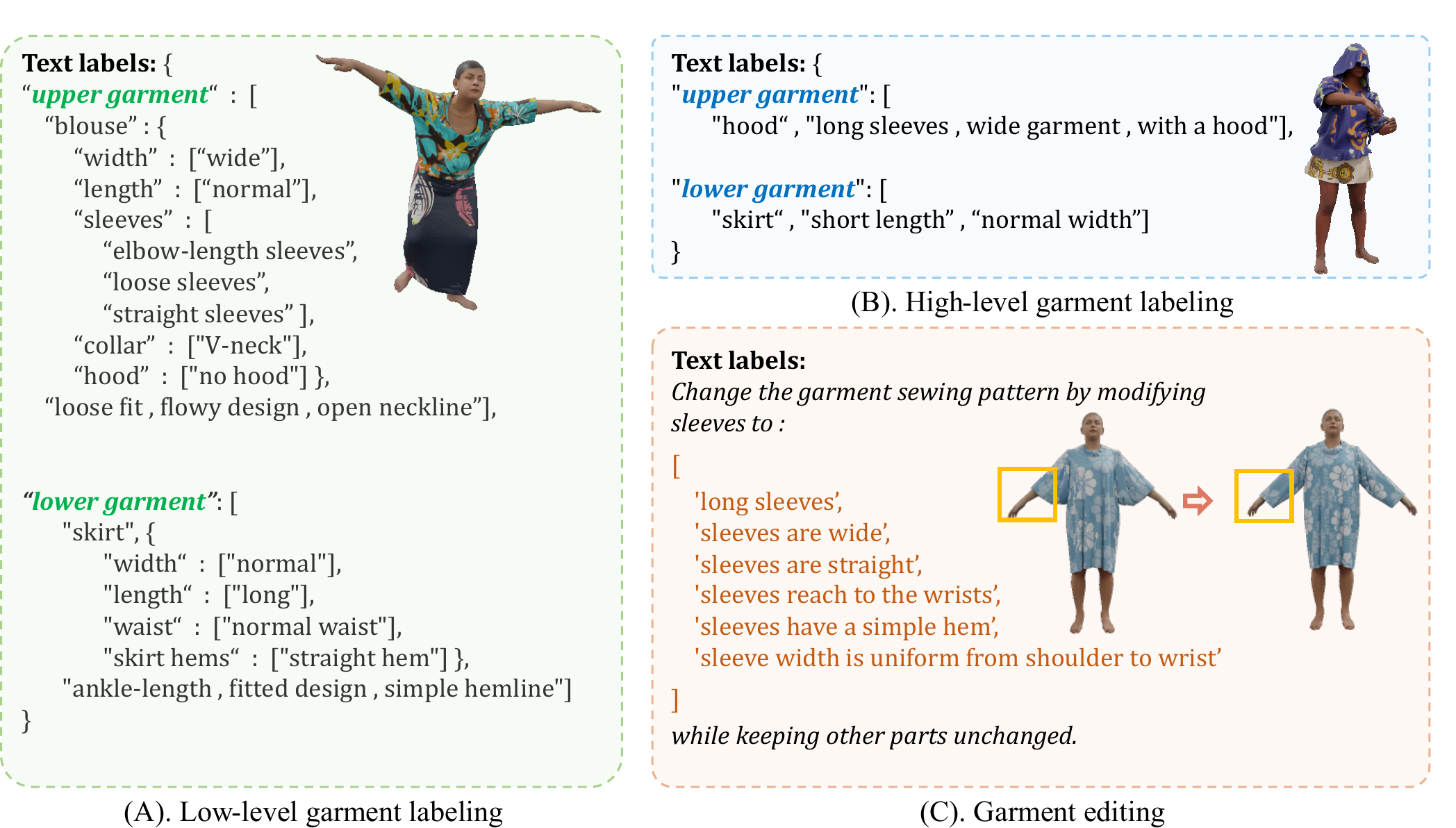}
    \captionof{figure}{Dataset samples include low-level, high-level, and garment editing descriptions.} 
    \label{fig:example_data}
\end{figure*}

\noindent{\textbf{Training Data Generation.}}
To create text descriptions for the garments in our training dataset, we render front and back images of the garments and then query GPT-4o to generate descriptions for the images. We use the prompts in~\cref{tab:gpt4o_image} to generate descriptions for each garment part, and use the prompts in~\cref{tab:gpt4o_wholeimage} to generate descriptions for all visible garment parts in the image. Additionally, we provide several examples from our dataset, including low-level and high-level garment descriptions, as well as garment editing descriptions; see in~\cref{fig:example_data}.

As described in~\cref{sec:method}, we construct question and answer pairs to finetune a multimodal LLM. Specifically, we build two datasets: an image-reconstruction dataset and a sewing pattern editing dataset. Text-based generation data is derived by removing the images from the image-reconstruction dataset. Detailed question lists of these datasets are illustrated in~\cref{tab:training_data_image_question,tab:training_data_text_image_question,tab:training_data_edit_question,tab:training_data_text_question} respectively. Example textual answers are shown in~\cref{tab:training_data_image_answers}, where \textcolor{Brown}{\texttt{[Sewing pattern without floats]}} refers to a JSON configuration where all float values are replaced with ``0''. Since the projection layer is used to calculate numeric values in the sewing pattern, it is unnecessary to output these numeric values directly in the textual answers. Replacing them with ``0'' simplifies the training process.
\begin{table}[h!]\centering
\begin{minipage}{\columnwidth}\vspace{0mm}    \centering
\begin{tcolorbox} 
    \centering
    \small
     \hspace{-6mm}
\begin{itemize}[leftmargin=2mm]
\footnotesize
\setlength{\itemsep}{1pt}
\item ``\textcolor{ForestGreen}{\texttt{<image>}} Can you estimate the outfit sewing pattern code in the image?"
\item ``\textcolor{ForestGreen}{\texttt{<image>}} Please estimate the outfit sewing pattern code."
\item ``\textcolor{ForestGreen}{\texttt{<image>}} Provide the sewing pattern codes for the garments according to the image."
\item ``\textcolor{ForestGreen}{\texttt{<image>}} What is the sewing pattern codes for the outfit shown in the image?"
\item ``\textcolor{ForestGreen}{\texttt{<image>}} Could you tell the outfit sewing pattern codes for the garments?"
\end{itemize}
\end{tcolorbox}
\caption{\textbf{Example questions for image reconstruction}. \textcolor{ForestGreen}{\textcolor{ForestGreen}{\texttt{<image>}}} is the placeholder token of the input image.}
    \label{tab:training_data_image_question}
\end{minipage}
\end{table}

\begin{table}[h!]\centering
\begin{minipage}{\columnwidth}\vspace{0mm}    \centering
\begin{tcolorbox} 
    \centering
    \small
     \hspace{-6mm}
\begin{itemize}[leftmargin=2mm]
\footnotesize
\setlength{\itemsep}{1pt}
\item ``\textcolor{ForestGreen}{\texttt{<image>}} Can you estimate the outfit sewing pattern code based on the image and the Json-format garment geometry description? \textcolor{Maroon}{\texttt{[Garment descriptions]}}"
\item ``\textcolor{ForestGreen}{\texttt{<image>}} Please estimate the outfit sewing pattern code based on the image and the garment geometry descriptions in Json format. \textcolor{Maroon}{\texttt{[Garment descriptions]}}"
\item ``\textcolor{ForestGreen}{\texttt{<image>}} Provide the sewing pattern codes for the garments according to the image and the Json-format garment geometry description. \textcolor{Maroon}{\texttt{[Garment descriptions]}}"
\item ``\textcolor{ForestGreen}{\texttt{<image>}} What is the sewing pattern codes for the outfit according to the image and the Json-format garment geometry description? \textcolor{Maroon}{\texttt{[Garment descriptions]}}"
\item ``\textcolor{ForestGreen}{\texttt{<image>}} Could you tell the outfit sewing pattern codes for the garments based on the image and the garment geometry descriptions in Json format? \textcolor{Maroon}{\texttt{[Garment descriptions]}}"
\end{itemize}
\end{tcolorbox}
\caption{\textbf{Example questions for text-guided image reconstruction.} \textcolor{Maroon}{\texttt{[Garment descriptions]}} refers to garment descriptions.}
    \label{tab:training_data_text_image_question}
\end{minipage}
\end{table}

\begin{table}[h!]\centering
\begin{minipage}{\columnwidth}\vspace{0mm}    \centering
\begin{tcolorbox} 
    \centering
    \small
     \hspace{-6mm}
\begin{itemize}[leftmargin=2mm]
\footnotesize
\setlength{\itemsep}{1pt}
\item ``Can you estimate the outfit sewing pattern code based on the Json-format garment geometry description? \textcolor{Maroon}{\texttt{[Garment descriptions]}}"
\item ``Please estimate the outfit sewing pattern code based on the garment geometry descriptions in Json format. \textcolor{Maroon}{\texttt{[Garment descriptions]}}"
\item ``Provide the sewing pattern codes for the garments according to the image and the Json-format garment geometry description. \textcolor{Maroon}{\texttt{[Garment descriptions]}}"
\item ``What is the sewing pattern codes for the outfit according to the Json-format garment geometry description? \textcolor{Maroon}{\texttt{[Garment descriptions]}}"
\item ``Could you tell the outfit sewing pattern codes for the garments based on the garment geometry descriptions in Json format? \textcolor{Maroon}{\texttt{[Garment descriptions]}}"
\end{itemize}
\end{tcolorbox}
\caption{\textbf{Example questions for text-based garment generation.} \textcolor{Maroon}{\texttt{[Garment descriptions]}} refers to garment descriptions.}
    \label{tab:training_data_text_question}
\end{minipage}
\end{table}

\begin{table}[h!]\centering
\begin{minipage}{\columnwidth}\vspace{0mm}    \centering
\begin{tcolorbox} 
    \centering
    \small
     \hspace{-6mm}
\begin{itemize}[leftmargin=2mm]
\setlength{\itemsep}{1pt}
\footnotesize
\item ``Adjust the old sewing pattern according to the text descriptions. The old garment sewing pattern is: \textcolor{RoyalBlue}{\texttt{[Old sewing pattern]}}. And the text descriptions are: \textcolor{Orchid}{\texttt{[Text descriptions]}}."
\item ``Adjust the old sewing pattern \textcolor{RoyalBlue}{\texttt{[Old sewing pattern]}} according to the text descriptions \textcolor{Orchid}{\texttt{[Text descriptions]}} without modifying other parts."
\item ``Here is an old garment sewing pattern: \textcolor{RoyalBlue}{\texttt{[Old sewing pattern]}}. Modify the pattern to align with the text descriptions: \textcolor{Orchid}{\texttt{[Text descriptions]}} without changing other parts."
\item ``Update the old sewing pattern following the text descriptions: \textcolor{Orchid}{\texttt{[Text descriptions]}}. The old garment sewing pattern is: \textcolor{RoyalBlue}{\texttt{[Old sewing pattern]}}."
\end{itemize}
\end{tcolorbox}
\caption{\textbf{Instructions for garment editing.}. \textcolor{RoyalBlue}{\texttt{[Old sewing pattern]}} refers to the initial garment sewing pattern to be edited, and \textcolor{Orchid}{\texttt{[Text descriptions]}} refers to the editing instructions.}
    \label{tab:training_data_edit_question}
\end{minipage}
\end{table}

\begin{table}[ht]\centering
\begin{minipage}{\columnwidth}\vspace{0mm}    \centering
\begin{tcolorbox} 
    \centering
    \small
     \hspace{-6mm}
\begin{itemize}[leftmargin=2mm]
\setlength{\itemsep}{1pt}
\footnotesize
\item ``\textcolor{Brown}{\texttt{[Sewing pattern without floats]}}."
\item ``It is \textcolor{Brown}{\texttt{[Sewing pattern without floats]}}."
\item ``Sure, it is \textcolor{Brown}{\texttt{[Sewing pattern without floats]}}."
\item ``The sewing pattern is \textcolor{Brown}{\texttt{[Sewing pattern without floats]}}."
\item ``The estimated sewing pattern is  \textcolor{Brown}{\texttt{[Sewing pattern without floats]}}."
\end{itemize}
\end{tcolorbox}
\caption{\textbf{Example textual answers for reconstruction, generation, and editing.} \textcolor{Brown}{\texttt{[Sewing pattern without floats]}} refers to a simplified sewing pattern in which numeric values are replaced with  ``0''. }
    \label{tab:training_data_image_answers}
\end{minipage}
\end{table}

\begin{table}[h!]\centering
\begin{minipage}{\linewidth}\vspace{0mm} 
\begin{tcolorbox} 
    \footnotesize
``I will provide an image of a human model wearing the \textcolor{Cyan}{\texttt{[garment name]}}. The top two subfigures show the front and back views of the model (from left to right), while the bottom two subfigures show the zoomed-in view of the front and back views of the \textcolor{Cyan}{\texttt{[garment name]}}. Please ONLY focus on the \textcolor{Plum}{\texttt{[part name]}} on the \textcolor{Cyan}{\texttt{[garment name]}}.\\

Please describe the geometries and structures of the \textcolor{Plum}{\texttt{[part name]}} on the \textcolor{Cyan}{\texttt{[garment name]}} according to the image. Strictly avoid mentioning other garment parts. Strictly avoid mentioning color, texture, seams, and material. \\

Return a Json LIST of several phrases, each describing a geometric feature of the \textcolor{Plum}{\texttt{[part name]}}, in the Json list format: [geometry feature 1, geometry feature 2, geometry feature 3, ...].''
\end{tcolorbox}
\caption{\textbf{GPT-4o prompts for generating garment part labels in an image.} \textcolor{Cyan}{\texttt{[garment name]}} refers to the name of the garment and \textcolor{Plum}{\texttt{[part name]}} refers to the name of the garment part.}
    \label{tab:gpt4o_image}
\vspace{-3mm}
\end{minipage}
\end{table}

\begin{table*}[h!]\centering
\begin{minipage}{\linewidth}\vspace{0mm} 
\begin{tcolorbox} 
    \footnotesize
I will provide one image of a human model wearing several garments. Describe the outer layer garments the models are wearing. In each image, the model may wear one upper garment and one lower garment, or the model may wear a single wholebody garment. Avoid describing extra accessories such as the scarves, socks, watch, badges, and etc. We have known that the model wears \textcolor{Cyan}{\texttt{[garment types]}}.\\

For each garment, you should generate THREE strings.\\

In the first string, describe the garment type (If THE SUBJECT HAS NAME, INCLUDE ITS NAME FIRST!);\\

  \quad Example phrases for the first string: "hood", "T-shirt", "jacket", "tuxedo", etc.\\

In the second string, describe the structures of the garment (DO NOT INCLUDE ANY INFO ABOUT THE HUMAN MODEL AND THE COLOR OF THE GARMENT) in the format of a dict. \\
    
    Select the keys from the following list: 
    ['width', 'length', 'sleeves', 'pant legs', 'waist', 'skirt hems', 'skirt hems', 'collar', 'hood', 'waist', ... ]\\
 
    In the value of the dict, please use several different short phrases in a list with the following tips: \\

        Describe the width of the garment: wide, normal, narrow, etc.\\
        \quad Describe the length of the garment: long, normal, short, etc.\\
        \quad Describe the length and width of the sleeves: long, normal, short, tight, loose sleeveless, etc.\\
        \quad Describe the detailed struture of the sleeves. Example: "asymmetrical sleeves", "straight sleeves", "puff sleeves", "three-quater sleeves", "accordion sleeves", etc.\\
        \quad Describe the length and width of the legs of trousers: long, normal, short, tight, loose legs, etc.\\
        \quad Describe the detailed struture of the pant legs. Example: "asymmetrical legs", "straight legs", "flared legs", "cropped legs", "cuffed legs", etc.\\
        \quad Describe the length and width of the dress: long, normal, short, tight, loose, etc.\\
        \quad Describe the detailed struture of the skirt hems. Example: "straight hem", "A-line hem", "pleated hem", "pencil hem", "slit hem", etc.\\
        \quad Describe the detailed struture of the neck or collar. Example: "crew neck", "V-neck", "turtle neck", "collarless", etc.\\
        \quad Describe the detailed struture of the hood. Example: "normal hood", "cape hood", "cowl hood", etc.\\

An example of the dict description for a T-shirt is: \\
{   
    'width': ['wide'],
    'length': ['normal'],
    'sleeves': ['elbow-length sleeves', 'tight sleeves', 'accordion sleeves'],
    'collar': ['crew neck'],
    'hood': ['no hood']
}\\

An example of the dict description for a skirt is: \\
{   
    'width': ['wide'],
    'length': ['knee-length'],
    'waist': ['high waist'],
    'skirt hems': ['pencil hem', 'pleated hem']
}\\

In the third string, describe the extra detailed structures of the garment (DO NOT INCLUDE ANY INFO ABOUT THE HUMAN MODEL AND THE COLOR OR PATTERN OF THE GARMENT) that are missing in the second string using several different short phrases split by ','. Example phrases for the third string: "pleated skirt", "high-waist", "zipper closure", "frayed hem", "mid-rise waist", etc.\\

Please strictly avoid mentioning color, texture, and material.\\

In the image, if the model is wearing one upper garment and one lower garment, return the results in the following format: {"upper garment": [upper garment type, upper garment geometric features, extra features], "lower garment": [lower garment type, lower garment geometric features, extra features]}. Otherwise, the model is wearing a single wholebody garment , return the results in the following format: {"wholebody garment": [wholebody garment type, wholebody garment geometric features, extra features]}. Only return the JSON dictionary in the above format with a length of 1 or 2.''
\end{tcolorbox}
\caption{\textbf{GPT-4o prompts for generating labels for all visible garment parts in an image.} \textcolor{Cyan}{\texttt{[garment types]}} refers to the types of the garments the model is wearing.}
    \label{tab:gpt4o_wholeimage}
\end{minipage}
\end{table*}

\begin{table}[h!]\centering
\begin{minipage}{\linewidth}\vspace{0mm} 
\begin{tcolorbox} 
    \footnotesize
    \begin{itemize}[leftmargin=2mm]
    \item ``\textcolor{ForestGreen}{\texttt{<image>}} Can you describe the garment outfits in in image in the Json format?''
    \item ``\textcolor{ForestGreen}{\texttt{<image>}} Can you estimate the outfit sewing pattern code based the image and the Json format garment geometry description?  \textcolor{Maroon}{\texttt{[Garment descriptions]}}''
    \end{itemize}
\end{tcolorbox}
\caption{\textbf{GPT-4o CoT prompts for generating sewing patterns from images.} \textcolor{Maroon}{\texttt{[Garment descriptions]}} refers to the textual descriptions of garments generated from the first question. }
    \label{tab:prompt_combine}
\end{minipage}
\end{table}

\begin{table}[h!]\centering
\begin{minipage}{\linewidth}\vspace{0mm} 
\begin{tcolorbox} 
    \footnotesize
    \begin{itemize}[leftmargin=2mm]
        \item ``\textcolor{ForestGreen}{\texttt{<image>}} Can you infer the garment material from the provided image input ? "
        \item ``\texttt{<existing material list>} Based on your inference, can you identify the material from the provided list that most closely matches the inferred physical properties ? "
    \end{itemize}
\end{tcolorbox}
\caption{\textbf{GPT-4o prompts for generating materials from images.} \texttt{<existing material list>} refers to the list of predefined materials. These prompts are used to initially infer the garment material from the existing material list and the provided image.
}
    \label{tab:prompt_params}
\end{minipage}
\end{table}

\subsection{Training Details}

We use LLaVA-1.5V-7B~\cite{liu2023llava} as our base VLM, integrating CLIP for vision encoding and Llama 2~\cite{touvron2023llama}, fine-tuned on conversational data, as the LLM backbone. We freeze the vision encoder and projection layers while finetune the LLM using LoRA~\cite{lora}. The sewing pattern projection layer is a two-layer (5120 x 76) MLP. The model is trained for 40 epochs, 500 steps per epoch, using the AdamW optimizer~\cite{kingma2014adam} with a learning rate of 1e-4. Training is done with a batch size of 4 per device on 4 NVIDIA H100 GPUs.

\subsection{Inference Details}
For image-based reconstruction, we apply the Chain-of-Thoughts (CoT)~\cite{wei2022chain} for \ours. Specifically, we first prompt ChatGarment to generate a detailed, JSON-format text description of the outfit in the image. The generated text descriptions are combined with the input image to estimate the final garment JSON configuration. Please see CoT prompts in~\cref{tab:prompt_combine}.

\subsection{Rule-based Simulation Control}\label{sec:simulation}

\ours demonstrates garment estimation and editing capabilities. The next step for artists is often simulating realistic garment movement. While existing tools need precise physical parameters to achieve desired deformations, we develop a rule-based method to derive material-specific parameters from text or images. This leverages LLM reasoning to map garment characteristics to simulation inputs.

We use C-IPC~\cite{Li2021CIPC} as the simulator because of its strong capability in dealing with complex interactions between the human body and garments. C-IPC requires several physical parameters like density, stretching stiffness, bending stiffness, and thickness, which are specific to the simulator and not directly derived from real-world measurements. To bridge this gap, we propose a hierarchical mapping approach. This involves initially matching inferred material properties to predefined material classes, followed by parameter refinement within each class.
For initialization, we prompt GPT-4o to identify the closest material match from a set of predefined material classes (see~\cref{tab:prompt_params}). The physics parameters for the target material are initially set based on the matched material, after which we further refine specific parameters that significantly impact the simulation behavior.

Our analysis demonstrates that four primary parameters, \textit{membE} (stretching stiffness), \textit{bendE} (bending stiffness), \textit{density}, and \textit{thickness}, show strong correlations with the high-level descriptors: \textit{rigid/soft}, \textit{heavy/light}, \textit{wrinkle/smooth}, and \textit{perceived thickness}. Moreover, LLM can effectively compare high-level material performance rather than directly estimating precise parameter values. Based on these correlations, each physical parameter is decoupled and individually mapped to its respective descriptor. We then ask GPT-4o to assign scores ranging from 1 to 10 for these high-level descriptors. These scores are used to adjust the corresponding physical parameters based on the score differences between the target material and the initial matched material, as described by the following equations:

\begin{align}
    \log\text{memb} &= \alpha_m \Delta_{\text{soft}} \cdot \log\text{memb}_{\text{base}} \\
    \log\text{bendE} &= \alpha_b \Delta_{\text{light}} \cdot \log\text{bendE}_{\text{base}} \\ 
    \text{density} &= \alpha_d \Delta_{\text{smooth}} \cdot \text{density}_{\text{base}} \\ 
    \text{thickness} &= \alpha_t \Delta_{\text{thickness}} \cdot \text{thickness}_{\text{base}} 
\end{align}
where $\Delta_{*}$ denotes the score differences derived from the inferred descriptors, allowing for refined adjustments of each parameter based on the closest matched material.

\section{Ablation Study Details}

\begin{table}
\tiny
\setlength{\tabcolsep}{3pt}
\renewcommand{\arraystretch}{1.3}
\centering
	\centering
     \resizebox{\columnwidth}{!}{
    \begin{tabular}{c|cc|cc}
       \multirow{2}{*}{Methods} & \multicolumn{2}{c|}{Dress4D} & \multicolumn{2}{c}{CLoSE} \\
        &  CD ($\downarrow$) & F-Score ($\uparrow$) & CD ($\downarrow$) & F-Score ($\uparrow$) \\
        \shline
        LLaVA-13B & 3.73 & \textbf{0.78} & \textbf{2.54} & 0.784 \\
        LLaVA-7B & \textbf{3.06} & \textbf{0.78} & {2.94} & \textbf{0.790} \\
    \end{tabular}}
\caption{\textbf{Ablation study: effect of multimodal LLM backbones.} Models utilizing LLaVA-7B and LLaVA-13B backbones demonstrate comparable performance on the two datasets.}
\label{table:ablation_model_size}
\vspace{-3mm}
\end{table}

\noindent\textbf{Multimodal LLM backbones.} 
As shown in \cref{table:ablation_model_size}, the LLaVA-7B and LLaVA-13B models achieve comparable results. For efficiency, we use the LLaVA-7B model for the other experiments in our paper.

\begin{table}
\tiny
\setlength{\tabcolsep}{3pt}
\renewcommand{\arraystretch}{1.3}
\centering
	\centering
     \resizebox{\columnwidth}{!}{
    \begin{tabular}{c|cc|cc}
       \multirow{2}{*}{Methods} & \multicolumn{2}{c|}{Dress4D} & \multicolumn{2}{c}{CLoSE} \\
        &  CD ($\downarrow$) & F-Score ($\uparrow$) & CD ($\downarrow$) & F-Score ($\uparrow$) \\
        \shline
        \ours* & 4.04 & \textbf{0.79} & 4.06 & 0.76 \\
        \ours & \textbf{3.06} & 0.78 & \textbf{2.94} & \textbf{0.79} \\
    \end{tabular}}
\caption{\textbf{Ablation analysis of different training datasets.} \ours* is only trained on high-level garment description datasets and exhibits poorer image reconstruction performance.}
\label{table:ablation_train_data}
\vspace{-3mm}
\end{table}

\noindent\textbf{Training Data.}
To assess the impact of part-level garment description datasets, we train a model (\oursnospace*) exclusively on general garment descriptions. For image-based reconstruction, we continue to use the Chain-of-Thoughts~\cite{wei2022chain} approach, prompting the model with a text description of the given garment as the first step.
As shown in~\cref{table:ablation_train_data}, the absence of part-level description datasets adversely affects image reconstruction results. In the Dress4D dataset ~\cite{wang20244d}, \oursnospace* exhibits a worse Chamfer distance but a slightly higher F-Score. In the CLoSE dataset~\cite{antic2024close}, \oursnospace* performs worse on both metrics.

\section{More Results}

\subsection{GarmentCodeRC}
\begin{figure}
    \captionsetup{type=figure}
    \centering
    \includegraphics[width=0.85\linewidth]{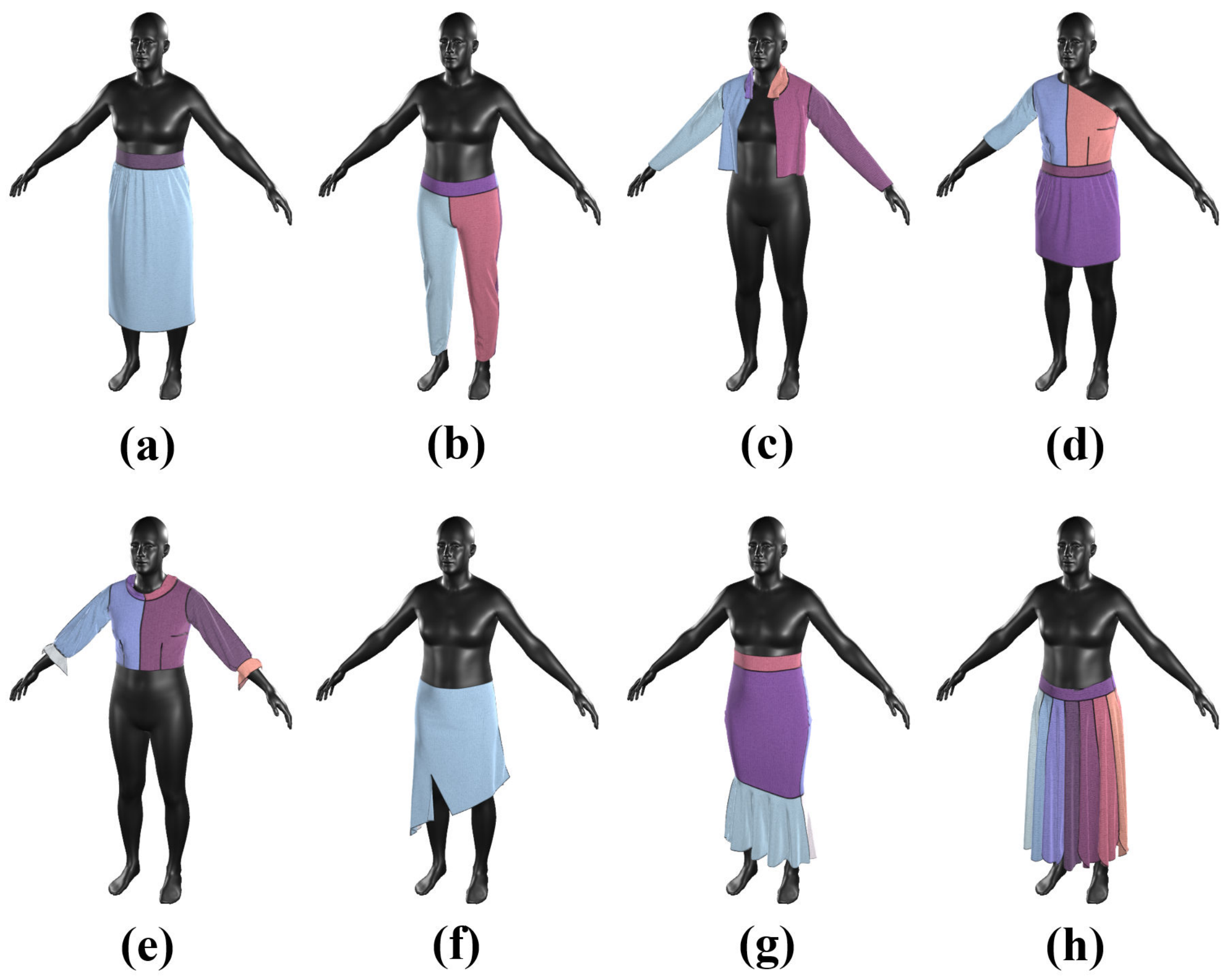}
    \captionof{figure}{\textbf{Examples of GarmentCodeRC garments.} The collection includes a high-waisted skirt (a), fitted pant legs (b), an open-front jacket (c), and various complex designs of dresses, shirts and skirts (d-h).}
    \label{fig: garmentcoderc_example}
\end{figure} 

{GarmentCode~\cite{GarmentCode2023,GarmentCodeData:2024} is an expressive DSL that can model complex garments with
geometric details, including various cuts, frills, and pleats. Built upon GarmentCode, our proposed GarmentCodeRC further enhances support for open-front jackets, high-waisted skirts, and fitted pant legs. Examples of GarmentCodeRC garments are shown in~\cref{fig: garmentcoderc_example}.}

\subsection{Text-based Generation}
\begin{figure*}
    \captionsetup{type=figure}
    \centering
    \includegraphics[width=0.86\linewidth]{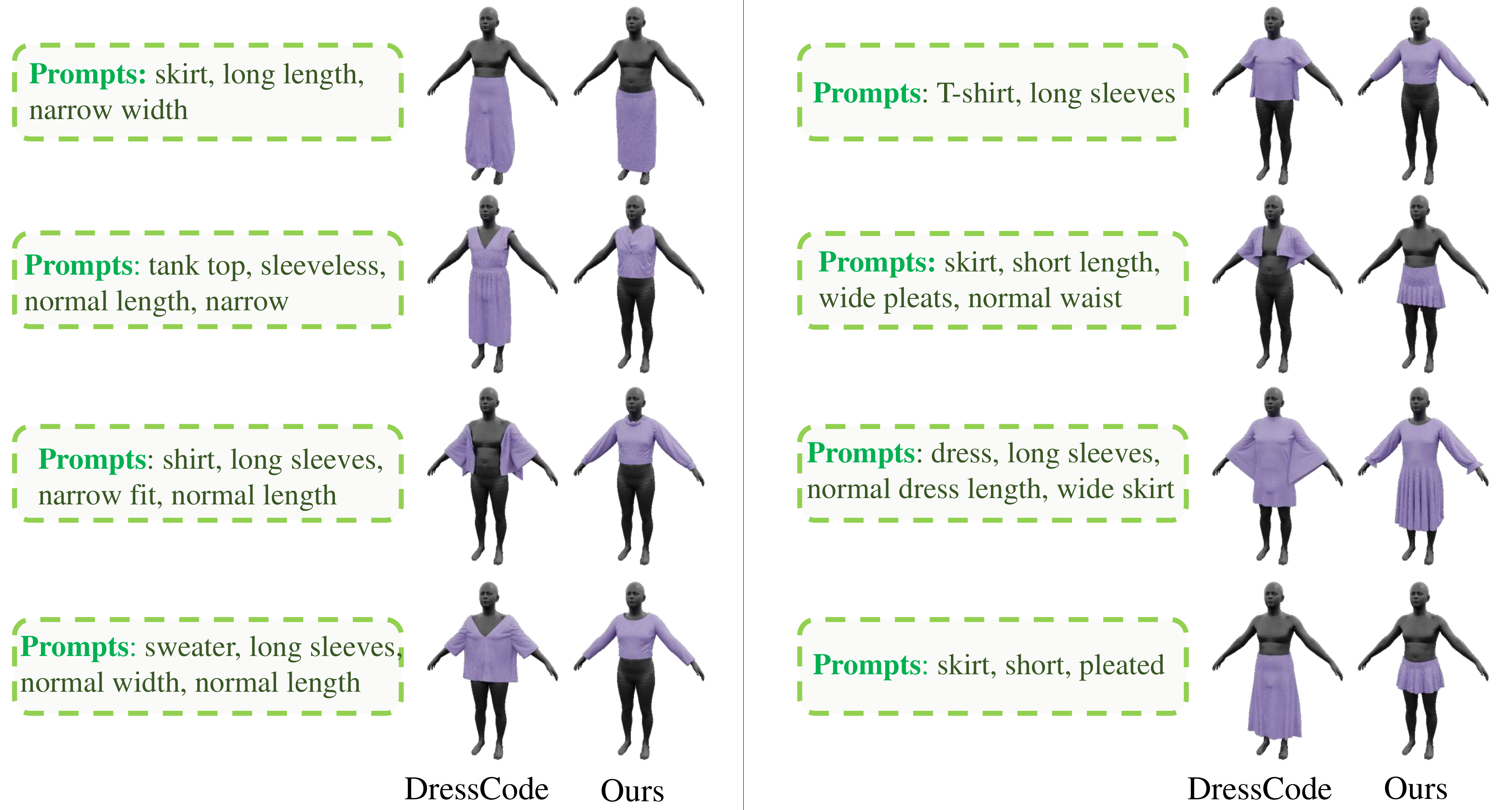}
    \captionof{figure}{\textbf{Text-based generation results.} \ours follows the instruction more accurately, generating more precise details (types, sleeves, length, etc.) compared to DressCode~\cite{he2024dresscode}.} 
    \label{fig: text_gen}
\end{figure*} 

\begin{figure*}[th]
    \captionsetup{type=figure}
    \centering
    \includegraphics[width=0.95\linewidth]{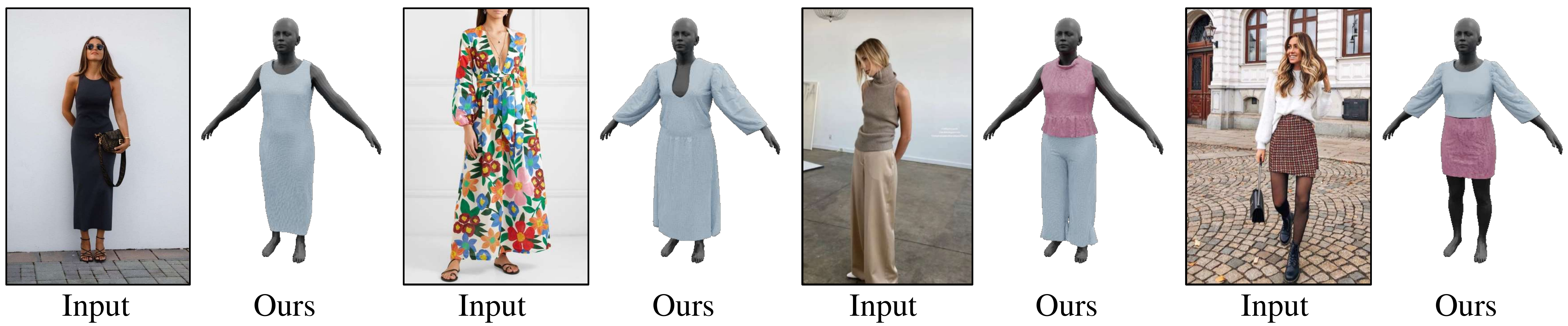}
    \captionof{figure}{\textbf{Single-turn Image-based Garment Reconstruction.} ChatGarment generates valid garments directly from the input images.}
    \label{fig: direct_image_recon}
\end{figure*} 

We provide qualitative examples of text-based garment reconstruction results in~\cref{fig: text_gen}, using the same prompt format as DressCode~\cite{he2024dresscode}. Compared to DressCode, \ours accurately generates garments with correct lengths, widths, and detailed features. In contrast, DressCode occasionally produces incorrect garment types, inaccurate sizes, and missing details.

\subsection{Single-turn Image-based Reconstruction}
{In our experiment, we apply the Chain-of-Thought~\cite{wei2022chain} method for optimized performance. However, ChatGarment also supports direct image-based reconstruction in a single-turn conversation. In this setup, ChatGarment is prompted to generate the garment JSON file directly from the input image. Qualitative examples are provided in~\cref{fig: direct_image_recon}.}

\subsection{Rule-based Simulation Control}
\begin{figure*}
    \captionsetup{type=figure}
    \centering
    \includegraphics[width=0.85\linewidth]{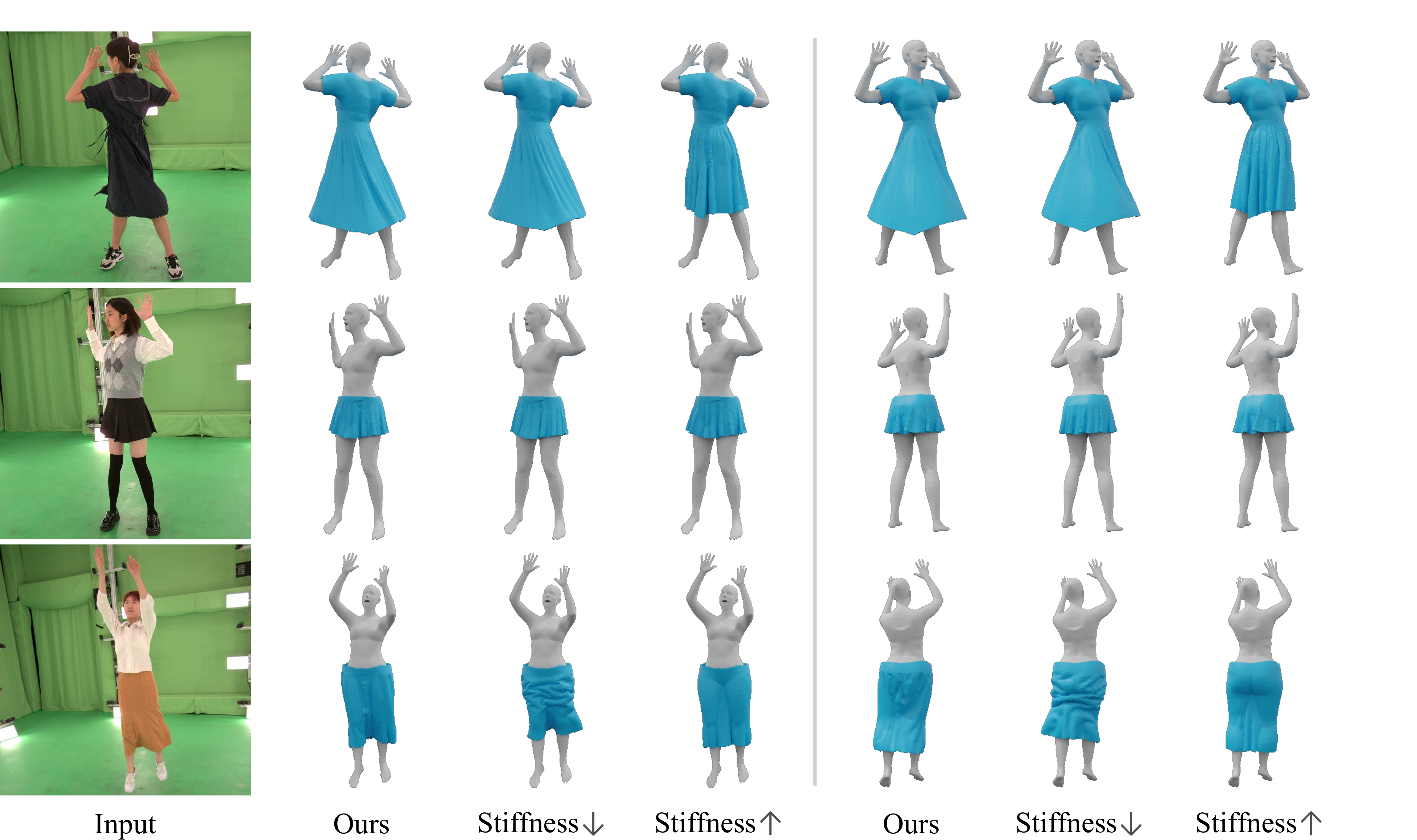}
    \captionof{figure}{\textbf{Rule-based Simulation Control.} We apply our rule-based method to estimate the simulation parameters corresponding to the input images. This approach also allows control over different physical deformation behaviors, such as those of soft materials like silk (Stiffness$\downarrow$) and rigid materials like denim (Stiffness$\uparrow$).}
    \label{fig: simulation_result}
\end{figure*} 

We present qualitative examples of rule-based simulation control in~\cref{fig: simulation_result}. The simulation parameters are aligned with the material characteristics in the input image as described in Sec.~\ref{sec:simulation}. Leveraging the high-level descriptors in our rule-based approach, we can also modify the simulation behavior to make the garment deform like other materials.
For instance, decreasing the stiffness (Stiffness$\downarrow$) results in a softer garment with more pronounced wrinkles and larger deformations under the same motion. Conversely, increasing the stiffness (Stiffness$\uparrow$) produces a garment with rigid material properties, making it less prone to stretching.

\subsection{Speed analysis of ChatGarment}
{We analyze garment reconstruction time on an A100 GPU. The process consists of three main stages: LLM decoding (12.1s), GarmentCode generation (3.5s), and sewing pattern stitching (33.9s). The primary bottleneck is the Warp-based sewing pattern stitching~\cite{warp2022,GarmentCode2023} stage. }

\section{Failure Cases and Future Work}
\begin{figure}
    \captionsetup{type=figure}
    \centering
    \vspace{-3mm}
\includegraphics[width=0.9\linewidth]{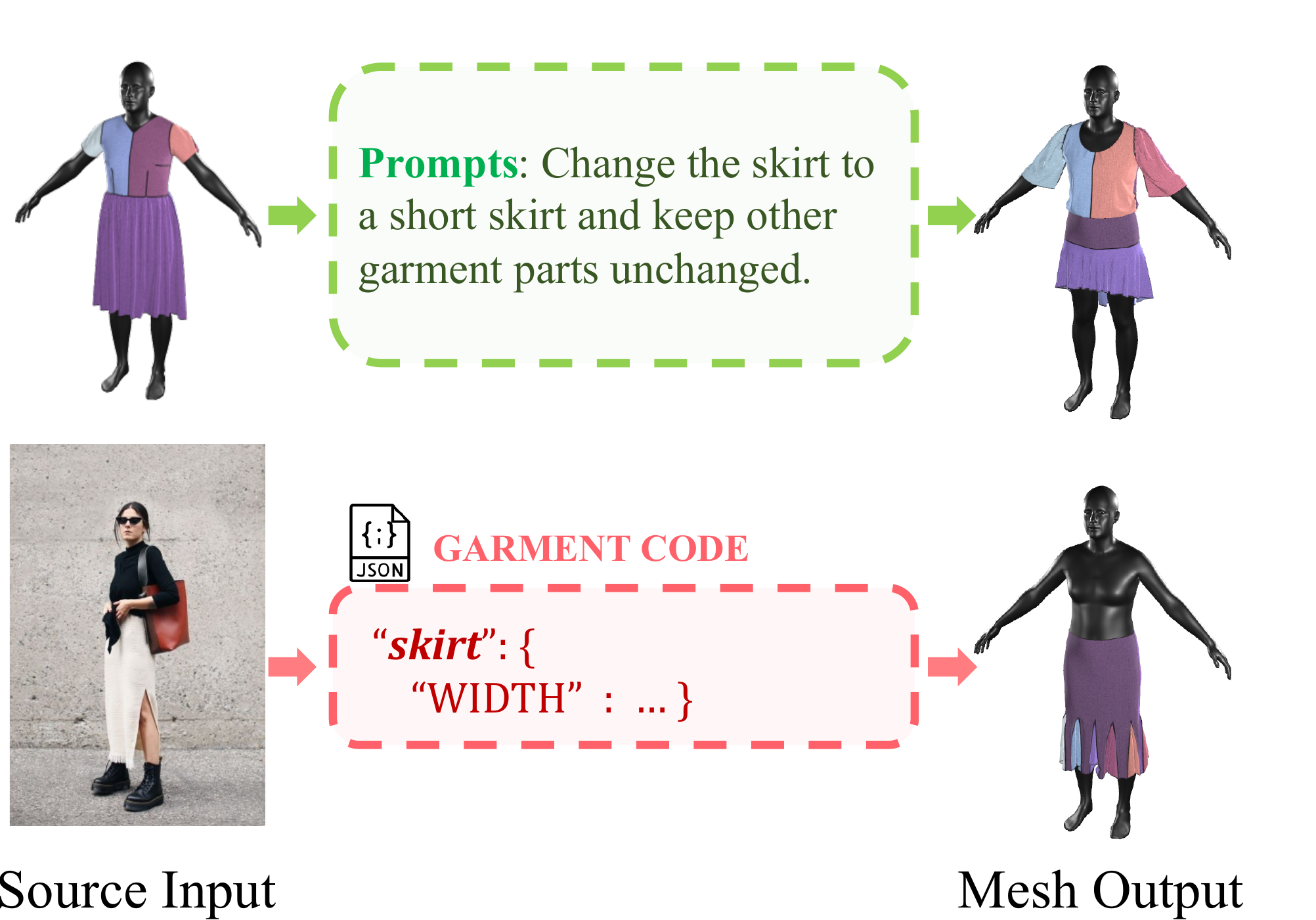}
    \captionof{figure}{\textbf{Failure cases of sewing pattern editing and reconstruction}. \ours might change the irrelevant garment parts (TOP: collars and sleeves). And \ours occasionally misinterprets the garment details (BOTTOM: skirt style).} 
    \label{fig: failure_edit}
\end{figure} 

{As shown in ~\cref{fig: failure_edit}, \ours occasionally struggles to edit specific garment parts without affecting other areas. For example, when adjusting the length of a skirt as requested, slight unintended changes may occur in the upper-body T-shirt. Additionally, in image-based garment reconstruction, it may fail to capture intricate details. While it can accurately identify the garment type as a skirt and estimate its length, it may misinterpret finer details, such as mistaking the bottom style of the skirt for pleats. These inaccuracies can be attributed to LLM hallucinations.
Additionally, although GarmentCodeRC can model complex garments with geometric details, including various cuts, frills, and pleats, as shown in ~\cref{fig: garmentcoderc_example}, it cannot model some specific details such as zippers and pockets.}

{Future improvements to our method could involve developing a more advanced programming parametric model for sewing patterns, which could enhance both the diversity of generated garments and the precision of garment editing. And hallucinations could be reduced via Retrieval-augmented Generation (RAG), in-context Learning (ICL), or LLM post-training. }

\end{document}